\pdfoutput=1 

\documentclass[journal]{IEEEtran}

\usepackage{graphicx}
\usepackage{algorithm}
\usepackage{algorithmic}
\usepackage{amsmath}
\usepackage{amsfonts}
\usepackage{subfigure}
\usepackage{multirow}
\usepackage{cite} 
\usepackage{caption}
\usepackage{color,xcolor}
\usepackage[colorlinks=true,
            linkcolor=black, 
            anchorcolor=black, 
            citecolor=black, 
            urlcolor=blue
            ]{hyperref}
\ifCLASSINFOpdf
\else
\fi
\begin{document}
%
\title{Semi-supervised Learning with Missing Values Imputation}
%
%
%

\author{Buliao~Huang, Yunhui~Zhu, Muhammad~Usman, Huanhuan~Chen,~\IEEEmembership{Senior Member,~IEEE}
\thanks{M. Shell was with the Department
of Electrical and Computer Engineering, Georgia Institute of Technology, Atlanta,
GA, 30332 USA e-mail: (see http://www.michaelshell.org/contact.html).}
\thanks{J. Doe and J. Doe are with Anonymous University.}
\thanks{Manuscript received April 19, 2005; revised August 26, 2015.}}

%
%

\markboth{Journal of \LaTeX\ Class Files,~Vol.~14, No.~8, August~2015}%
{Shell \MakeLowercase{\textit{et al.}}: Bare Demo of IEEEtran.cls for IEEE Journals}
%



\maketitle

\begin{abstract}

Incomplete instances with various missing attributes in many real-world applications have brought challenges to the classification tasks. Missing values imputation methods are often employed to replace the missing values with substitute values. However, this process often separates the imputation and classification, which may lead to inferior performance since label information are often ignored during imputation. Moreover, traditional methods may rely on improper assumptions to initialize the missing values, whereas the unreliability of such initialization might lead to inferior performance. To address these problems, a novel semi-supervised conditional normalizing flow (SSCFlow) is proposed in this paper. SSCFlow explicitly utilizes the label information to facilitate the imputation and classification simultaneously by estimating the conditional distribution of incomplete instances with a novel semi-supervised normalizing flow. Moreover, SSCFlow treats the initialized missing values as corrupted initial imputation and iteratively reconstructs their latent representations with an overcomplete denoising autoencoder to approximate their true conditional distribution. Experiments on real-world datasets demonstrate the robustness and effectiveness of the proposed algorithm.

\end{abstract}


%
\IEEEpeerreviewmaketitle

\section{Introduction}
\IEEEPARstart{M}{issing} value problem is ubiquitous in real-world applications  \cite{wang2019industrial}. Inappropriate treatment for incomplete instances with missing values might consequently degrade the performance of machine learning algorithm. Therefore, how to appropriately handle missing values in classification problems is still a challenge for the machine learning community.

In order to handle missing values in classification, one straightforward approach is to employ a “two-step” strategy, which first applies unsupervised missing value imputation methods \cite{Strawderman89,2014Mice,stekhoven2011missforest,9157706,GAIN} to replace the missing values with substitute values, and subsequently employs supervised classification methods on the imputed dataset\footnote{The unsupervised missing value imputation methods are often trained on both labelled and unlabelled incomplete instances to impute their missing values, while the supervised classification methods are often trained on the labelled imputed instances to predict the unobserved labels of unlabelled imputed instances.}. Recently, deep generative methods, such as normalizing flow \cite{dinh2015nice,dinh2017density,10.5555/3327546.3327685}, have led to great advances in unsupervised missing value imputation methods attributable to their powerful distribution estimation and minor dependence on distribution assumptions\footnote{Normalizing flow-based model could express any distribution  under reasonable conditions. The details of the proof and the conditions could be viewed in \cite{abs-1912-02762}. }. For example, Richardson et al. \cite{9157706} proposed Monte Carlo Flow (MCFlow), which leverages normalizing flow to estimate the data distribution of the incomplete instances and imputes the missing values by Monte Carlo sampling from the estimated distribution. However, the separation between imputation and classification in these “two-step” methods may lead to inferior performance since label information are often ignored during imputation.  

\begin{figure}[t]
\centering
\begin{minipage}[c]{0.48\columnwidth}
\subfigure[ imputation with estimated data distribution]{
    \label{fig:subfig:b} 
	\includegraphics[width=0.9\textwidth]{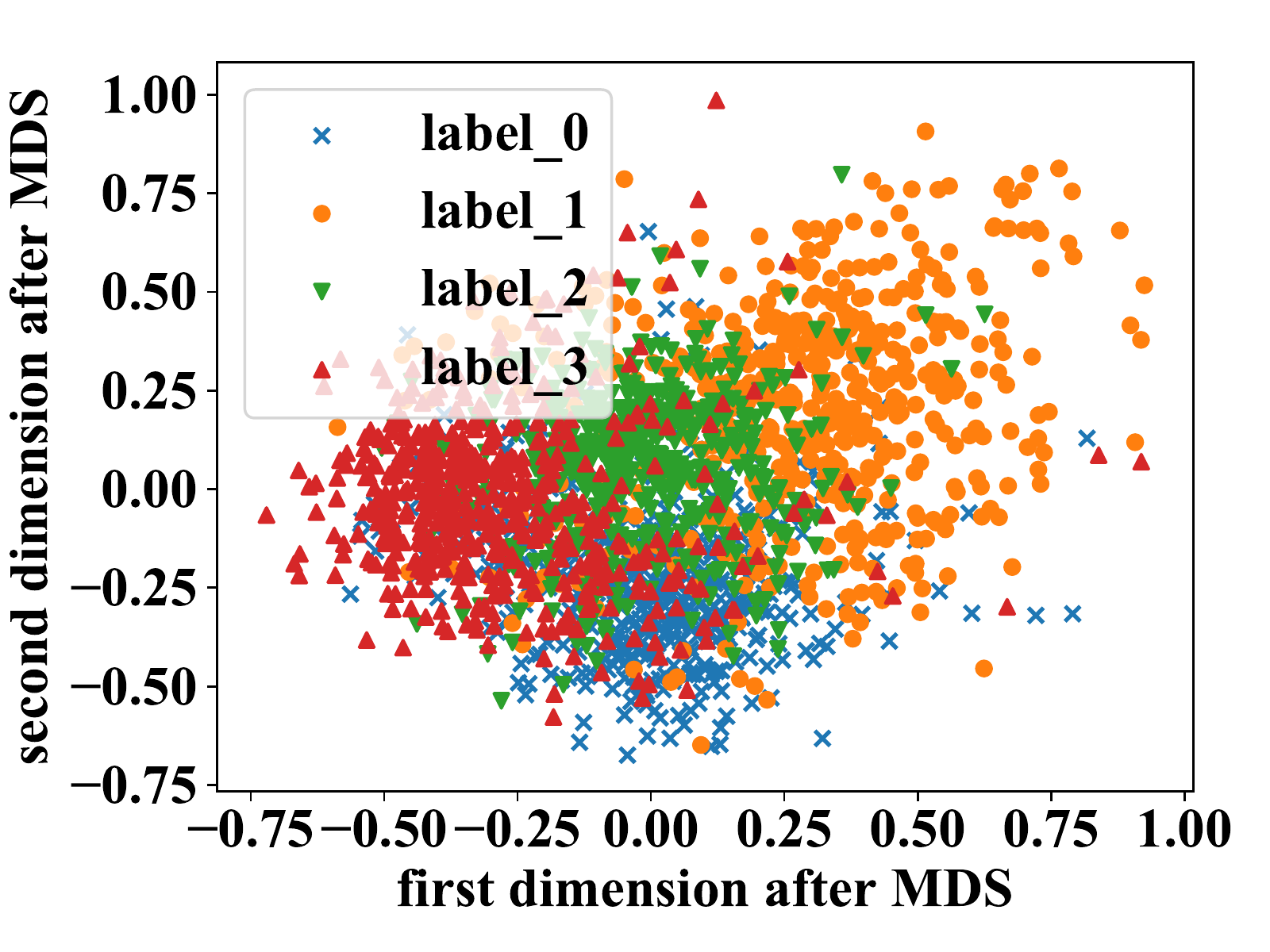} }
\end{minipage}
\begin{minipage}[c]{0.48\columnwidth}
\subfigure[ imputation with estimated \textbf{conditional distribution}]{
    \label{fig:subfig:b} 
	\includegraphics[width=0.9\textwidth]{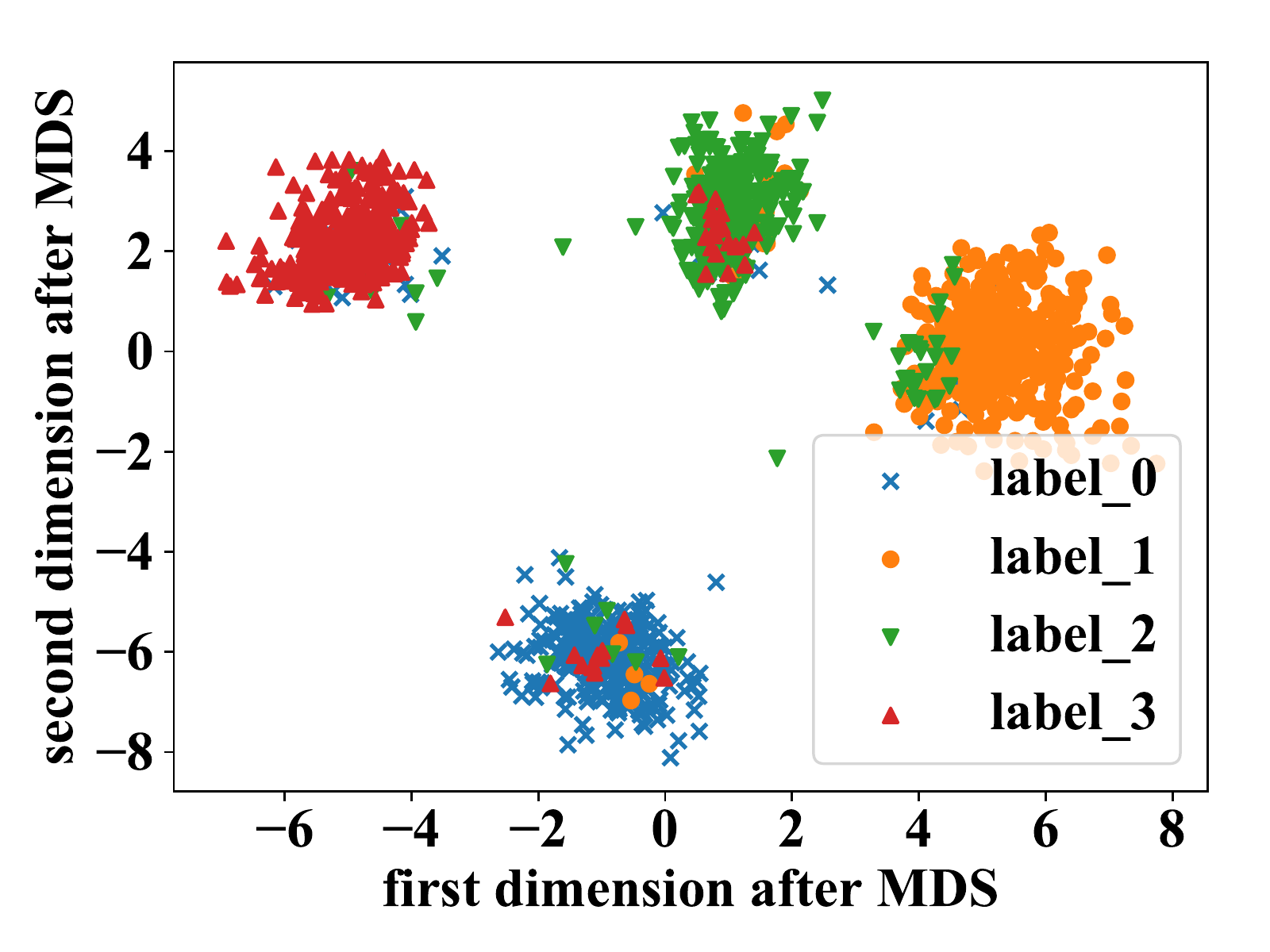} }
\end{minipage}
\caption{Examples\protect\footnotemark \  for the imputation result with estimated data distribution and \textbf{conditional distribution} from the UCI dataset \emph{wifi localization} (Detailed settings could be viewed in Sec. \ref{Comparison}).  It's obvious that imputation with estimated conditional distribution could be classified with better interpretability.}
\label{unified advantage}
\end{figure}
\footnotetext{Multi-dimensional scaling (MDS) \cite{kruskal1978multidimensional} is employed to reduce the dimensionality to 2D for better visualization.}

\begin{figure*}[t]
\centering
\includegraphics[width=1.95\columnwidth]{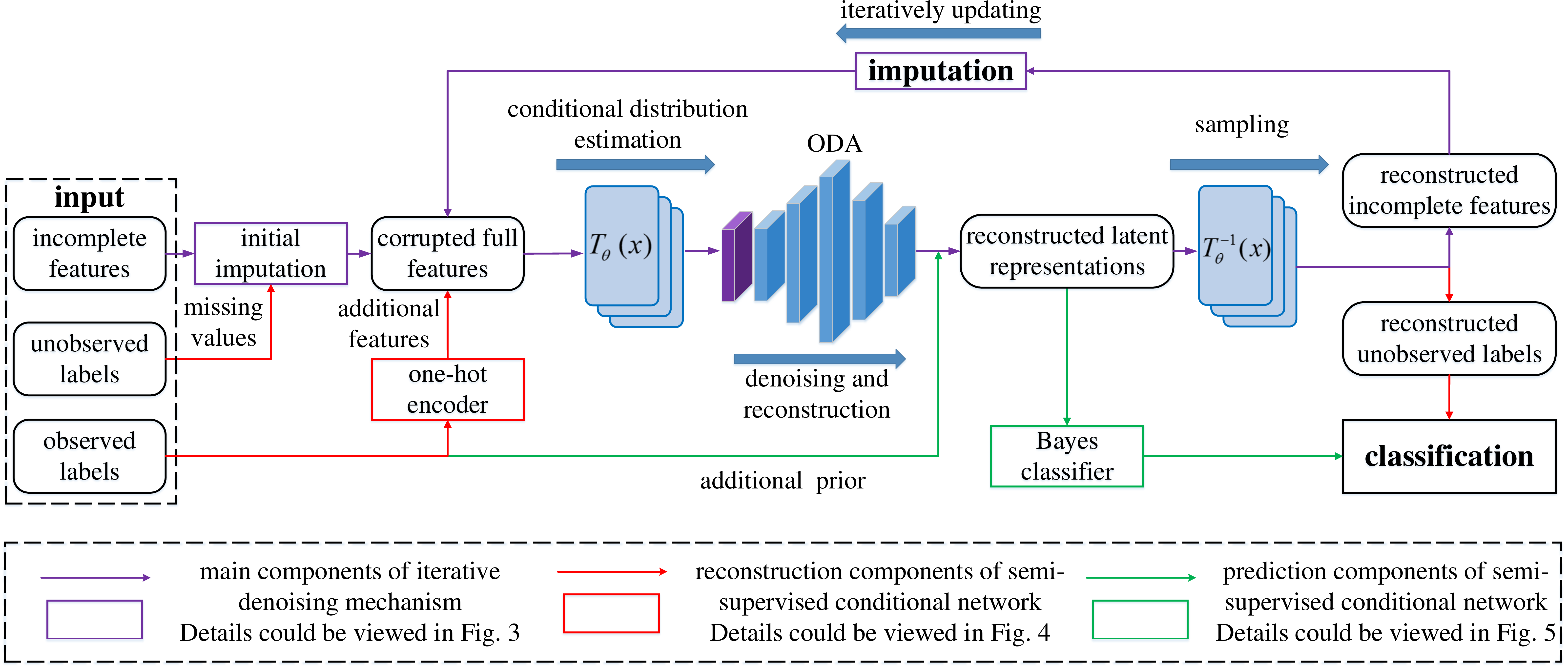} 
\caption{The architecture of the proposed SSCFlow, where normalizing flow $T_\theta(x)$ estimates the conditional distribution of incomplete instances for imputation and classification. SSCFlow explicitly takes the observed labels as additional features and prior knowledges to facilitate the conditional distribution estimation in $T_\theta(x)$. The unobserved labels are taken as missing values to directly participate in the conditional distribution estimation, and are reconstructed (predicted) to accomplish classification. Meanwhile, SSCFlow iteratively reconstructs the corrupted full features to approximate the true conditional probability density of missing values for imputation. }
\label{whole_structure}
\end{figure*}

The aforementioned problem could be addressed by replacing the data distribution estimation in MCFlow \cite{9157706} with \textbf{conditional distribution} estimation conditioned on the labels of incomplete instances. The conditional distribution estimation could facilitate the imputation by utilizing label information to introduce additional controllability in sampling \cite{MirzaO14,10.1007/978-3-030-69538-5_37,8924906}. Furthermore, as illustrated in Fig. \ref{unified advantage}, the classification could be accomplished with better interpretability by modelling Bayes decision rule with conditional distribution \cite{izmailov2019semisupervised}. Integrating these two concepts, this combines the unsupervised imputation task and supervised classification task into a single semi-supervised task. This new task requires simultaneous conditional distribution estimation of both labelled and unlabelled incomplete instances for their imputation and classification tasks. However, it remains a challenging problem to estimate the conditional distribution of unlabelled incomplete instances in existing semi-supervised deep generative methods \cite{izmailov2019semisupervised,trippe2018,atanov2020semiconditional}. Firstly, it may be hard to directly estimate the conditional distribution of \textbf{unlabelled} instances due to their unobserved labels. Therefore, existing methods often estimate it indirectly by assuming that it is consistent with the conditional distribution of the labelled instances and estimating the latter. However, such indirect estimation may ignore the latent complex dependencies between the conditional distribution of unlabelled instances and their unobserved labels, which may lead to biased results. Furthermore, most of existing methods assume the dataset to be completely observed, resulting in inability to handle \textbf{incomplete} input. Therefore, they often need to initialize the missing values, whereas the unreliability of such initialization might lead to inferior performance and improper assumptions on the missing values may be made \cite{YouMDKL20}.

To tackle these challenges, Semi-supervised Conditional Normalizing Flow (SSCFlow) with two novel algorithms is presented in this paper. Firstly, SSCFlow employs a novel semi-supervised conditional network, wherein the structure is shared between labelled and unlabelled incomplete instances to directly estimate their conditional distribution. In the semi-supervised conditional network, the observed labels are taken as additional input features and prior knowledges to guide the conditional distribution estimation. As for the unobserved labels, they are treated as missing values to directly participate in the conditional distribution estimation, and are reconstructed (predicted)\footnote{From the view of imputation task, the unobserved labels are reconstructed. From the view of classification task, the unobserved labels are predicted.} by maximizing the joint likelihood over labelled and unlabelled instances. Secondly, a novel iterative denoising mechanism is proposed in SSCFlow to tackle the unreliable initialization of missing values. The iterative denoising mechanism takes the initialized missing values as corrupted initial imputation and iteratively reconstructs their latent representations\footnote{Normalizing flows estimate probability density by transforming the data to its latent representation with invertible transformations and computing the relative change of local density volume \cite{9089305}. Details could be viewed in Sec. \ref{background}.} with an overcomplete denoising autoencoder (ODA) to approximate the true conditional distribution of incomplete instances.

In summary, this paper makes the following contributions: 
\begin{itemize}
\item This paper proposes the novel semi-supervised conditional normalizing flow (SSCFlow). To the best of our knowledge, SSCFlow is the first method that estimates the conditional distribution of incomplete instances to accomplish their imputation and classification simultaneously.
\item A novel semi-supervised conditional network is proposed, wherein the structure is shared between labelled and unlabelled incomplete instances to facilitate their imputation and classification by directly estimating their conditional distribution.
\item A novel iterative denoising mechanism is proposed to handle the incomplete input and make SSCFlow robust against the unreliability of the initialization for missing values.
\item The Bayes decision rule based classification in the proposed method naturally encodes clustering principle and has strong interpretability.
\end{itemize}

The rest of this paper is organized as follows. The background of the normalizing flow, which is the base model of SSCFlow, is illustrated in Sec. \ref{background}. Afterwards, a review of the related work is provided in Sec. \ref{related}. In Sec. \ref{Description}, the task of imputation and classification with missing values is described. Sec. \ref{methods} introduces the proposed methods in detail. Sec. \ref{expr} details the experimental settings and empirical analysis. Finally, the conclusion and future work are provided in Sec. \ref{conclusion}.

\section{Background: Normalizing Flow}
\label{background}

SSCFlow builds on normalizing flow \cite{dinh2015nice,dinh2017density,10.5555/3327546.3327685}, a type of deep generative model as variational auto-encoder (VAE) \cite{KingmaW13} and generative adversarial nets (GAN) \cite{2014Generative}. Compared with VAE and GAN, normalizing flow has highly desirable properties like exact density estimation and exact latent-variable inference. Normalizing flow models data distribution by transforming a prior distribution  (e.g., a standard normal \cite{9089305}) in the latent space into a more complex distribution in the input space with a sequence of invertible and differentiable transformations. Let us assume the task to find a join distribution over $x$, where $x$ is a D-dimensional real vector. The main idea of flow-based methods is to express $x$ as a sequence of transformations $T^{-1}$ of a latent representation vector $z$ sampled from a latent prior distribution $P_Z(z)$.
\begin{flalign*}
\label{Normalizing flows}
&& x=T^{-1}_\theta(z)  \ where \ z\sim P_Z(z) &&\refstepcounter{equation}\tag{\theequation}
\end{flalign*} where $T^{-1}$ is parameterized by a set of parameters $\theta$. According to theory of flow-based methods, the transformations $T^{-1}_\theta$ must be invertible and both $T_\theta$ and $T_\theta^{-1}$ must be differentiable.  Under these conditions, the density of $x$ could be evaluated by transforming it back to the latent prior distribution and computing change of variables \cite{MT} as following :   \begin{flalign*}
\label{change of variable theorem}
&& P_X(x) = P_Z(T_\theta(x))|det J_{T_\theta}(x)| &&\refstepcounter{equation}\tag{\theequation}
\end{flalign*} where $J_{T_\theta}$ refers to the Jacobian of $T_\theta$. Since flow models give the exact density likelihood in Eq. \ref{change of variable theorem}, $\theta$ can be trained by directly optimizing the log likelihood \cite{li2020flow} as following: \begin{flalign*}
\label{train log likelihood}
&& \theta^* &= \mathop{\arg\max}_{\theta} P_X(x) \\
&& &= \mathop{\arg\max}_{\theta} P_Z(T_\theta(x))|det J_{T_\theta}(x)|\\
&& &= \mathop{\arg\max}_{\theta} log(P_Z(T_\theta(x)) + log(|det J_{T_\theta}(x)|). &&\refstepcounter{equation}\tag{\theequation}
\end{flalign*} In addition, owing to the invertibility of the transformations, one can draw samples by simply inverting the transformations over a set of samples from the latent space.




\section{Related Work}
\label{related}
\begin{figure*}[t]
\centering
\includegraphics[width=2\columnwidth]{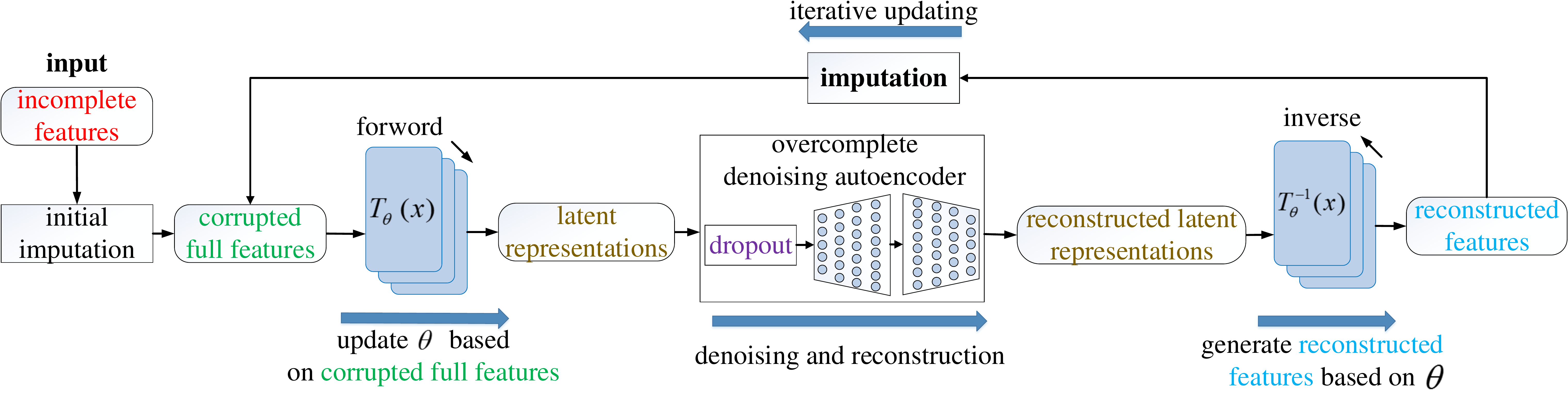} 
\caption{High-level view of iterative denoising mechanism. To tackle the incomplete input, random initialization is employed as initial imputation to generate corrupted full features. The iterative denoising mechanism takes the unreliability of the initialization into account and iteratively reconstructs the latent representations of the corrupted full features with an overcomplete denoising autoencoder to approximate the true distribution of missing values for imputation.  }
\label{IDM}
\end{figure*}

In this section, we briefly summarize existing literature for: methods for classification with missing values, learning in model space, and semi-supervised normalizing flow.
\subsection{Methods for Classification with Missing Values}
A handful number of imputation methods \cite{2014Mice,9157706,GAIN,6412787,6979248,7959549,8332950,8418828,9370000,9407339,9418555,9457222} have been proposed to replace the missing values with estimated values as a preprocessing step for the classification task. For example, Hapfelmeier et al. \cite{2014Mice} proposed the multivariate imputation by chained equations (MICE) to suggest multiple imputations (replacement values) for multivariate missing data. The method is based on Fully Conditional Specification, where each incomplete variable is imputed by a separate model. Stekhoven et al. \cite{stekhoven2011missforest} proposed missForest, a non parametric and mixed-type imputation method applicable to all data types. By averaging over many unpruned classification or regression trees, missForest intrinsically constitutes a multiple imputation scheme, using the built-in out-of-bag error estimates of random forest. Recently, deep generative models have led to great advances in unsupervised missing value imputation. Greg et al. \cite{GAIN} proposed Generative Adversarial Imputation Nets (GAIN) to impute missing data by adapting the well-known Generative Adversarial Nets (GAN) framework. Q. Shi et al. \cite{9157706} proposed MCFlow, which leverages normalizing flow generative models and Monte Carlo sampling for imputation.

There are also some methods \cite{10.1145/2939672.2939785,6848791,8491306} that could directly classify the incomplete datasets without imputation. However, they may ignore the potential contribution of these missing values for classification, which may lead to inferior classification performance. In addition, the dataset remains incomplete once the classification task is completed, which may obstruct other sub-sequent tasks. Due to the issues pointed above, this paper focuses simultaneous imputation and classification for incomplete instances with missing values.

\subsection{Learning in the Model Space}
The proposed iterative denoising mechanism could be viewed as a model to be fitted on the latent representations of the incomplete instances in the latent space to have more stable and parsimonious representations of them to estimate their distribution. This technique coincides with the new trend, termed ``learning in the model space" in the machine learning community.  The  idea  of  learning  in  the  model  space (LiMS) \cite{chen2013learning}  is  to  use  models fitted  on parts of data  as their  stable   and  parsimonious representations. Afterwards,  more robust and more targeted learning on diverse data collections could be achieved by directly performing learning in  the  model  space  instead  of  the  original  data  space. LiMS origins from time series learning \cite{10.1145/2487575.2487700} and fault diagnosis \cite{2014Cognitive}, and it has been improved over time by considering both representation ability of models, and the discrimination ability in the model space  \cite{chen2015model}. More recently, LiMS has been extended to symbolic sequence learning  \cite{article}, multi-objective learning \cite{2018Multiobjective}, imbalanced learning  \cite{2016Model}, dynamic state modelling  \cite{2017Sequential}, and short sequence learning \cite{2019Short}, etc.

\subsection{Semi-Supervised Normalizing Flow}
Normalizing flows have been used in the past to estimate the conditional distribution of unlabelled instances in semi-supervised learning. For instance, Trippe et al. \cite{trippe2018} propose to use distinct flows for each class to condition unlabelled instances on the class labels. However, it doesn't share weights between classes and may lead to ineffective utilization of training data \cite{atanov2020semiconditional}. To fix this issue, Atanov et al. \cite{atanov2020semiconditional} propose a conditional coupling layer with a multi-scale architecture. Although this approach shares weights between classes to a certain degree, the conditional coupling layer still requires linear time w.r.t. number of classes. Izmailov et al. \cite{izmailov2019semisupervised} proposed FlowGMM, which models the density in the latent space as a Gaussian mixture with each mixture component corresponding to a class. However, since the label information only participates in the loss computation and does not affect the generative process, it yields limited improvements for imputation. In addition, the parameters of its Gaussian mixture are fixed after random initialization, which may lead to unstable convergence. It is pertinent to note that none of these approaches could handle incomplete input.

\section{Problem Description}
\label{Description}


Let us assume that there is an incomplete dataset $X = \{x_0,x_1,...,x_{n-1}\}$ with $n$ instances. The $i$-th instance $x_i$ in $X$ contains $d$ attribute values and can also be written as $( x_i^1 , x_i^2 , ... , x_i^d )$. A binary masking vector $M(x_i) \in (0,1)^{d}$ is assigned to $x_i$ to indicate the location of missing attributes with corresponding masking attributes set to $0$. Generally, $X$ could be divided into two subset as training dataset $X_{tr}$ with label set $Y_{tr} \in {1,...,L}$ observed to indicate its labels ($L$ classes) and test dataset $X_{te}$ with their unobserved labels $Y_{te}$.  The task of this paper is to assign each missing value in the incomplete dataset $X$ with its substituted value and predict the unobserved labels $Y_{te}$. Considering that both the training dataset $X_{tr}$ and test dataset $X_{te}$ may contain missing values that need to be imputed, this paper forms the task in a semi-supervised manner and takes the instances in the test dataset as unlabelled instances when they are used in the following algorithms.

\section{Semi-supervised Conditional Normalizing Flow}
\label{methods}

In this section, the SSCFlow model is introduced. Given an incomplete instance, SSCFlow imputes its missing values and predicts its class label simultaneously. Firstly, the normalizing flow with a novel iterative denoising mechanism is introduced in a way that imputation doesn't depend upon unreliable initializations. Afterwards, a novel semi-supervised conditional network is proposed, wherein the structure is shared between labelled and unlabelled incomplete instances to facilitate their imputation and classification by directly estimating their conditional distribution. Finally, the training process of SSCFlow with novel loss functions is illustrated. The architecture of the proposed SSCFlow is given in Fig. \ref{whole_structure}.



\subsection{Iterative Denoising Mechanism}
\begin{figure*}[t]
\centering
\includegraphics[width=2\columnwidth]{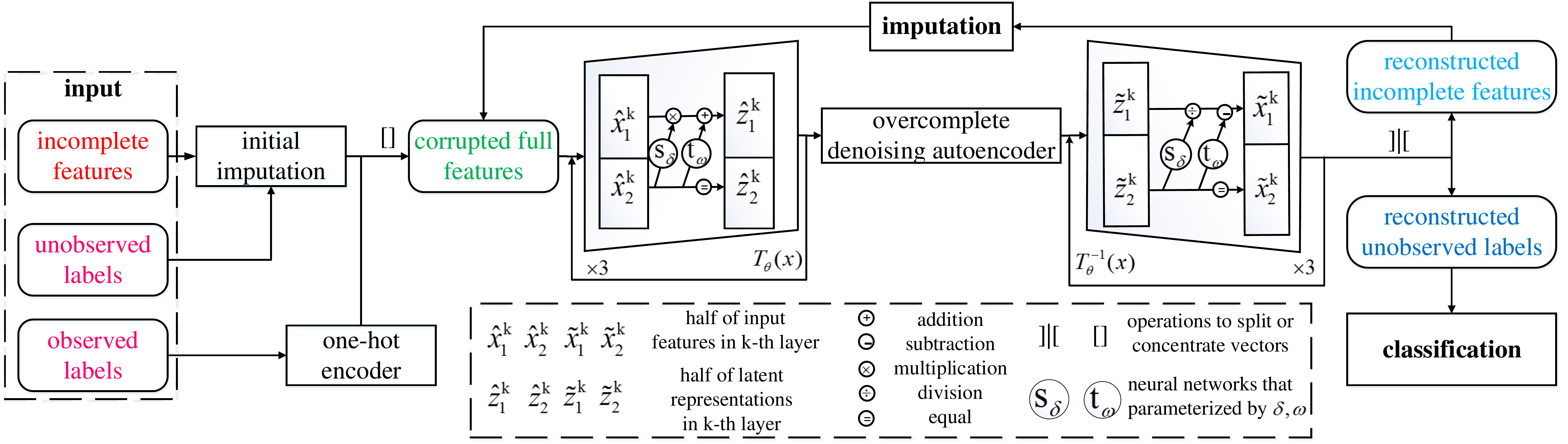} 
\caption{High-level view of semi-supervised conditional invertible transformation. The observed labels are taken as additional features to direct the conditional distribution estimation, while the unobserved labels are taken as missing values and reconstructed during imputation.}
\label{affine coupling layer}
\end{figure*}
Since normalizing flow is a generative model, data imputation can be performed by sampling from the underlying data distributions learned by the model. Assuming that there is a completely observed subset of training data $X_{tr}$ that could optimize the parameters $\theta$ of the invertible transformation $T_\theta$ in Sec. \ref{background} to learn the data distribution of completely observed subset, the missing values of the incomplete subset could be imputed by sampling from their marginal distribution based on their observed values and the trained normalizing flow $T_\theta$. However, since the training data may also have missing entries, the direct estimation of data distribution is not possible for incomplete data, posing a challenge to the training of normalizing flow.


Following MCFlow \cite{9157706}, SSCFlow addresses this challenge with an alternating algorithm. Specifically, an initialization is first employed to fill the incomplete instances $x$ as imputed instances $\dot{x}$. Afterwards, SSCFlow iteratively updates the generative network parameters $\theta$ according to the imputed instances $\dot{x}$ and updates the imputation of $\dot{x}$ according to the updated $\theta$. In MCFlow, a multi-layer perceptron (MLP) with fixed hidden units in its each layer is employed in the latent space to find the latent representation $\dot{z}$ of $\dot{x}$ with the maximum density likelihood that maps to
a data vector whose entries match the observed values.

It is pertinent to note that the unreliability of the initialization has not been considered in MCFlow \cite{9157706} and improper assumptions about the missing values may be made by initializing them with special default values \cite{YouMDKL20}. Moreover, since the MLP only requires the entries of the resultant vector to match the observed values, it easily overfits the observed values. Therefore, MCFlow only tries to fit on the marginal distribution of the observed values rather than learning the data distribution of the unobserved missing values. To resolve this issue, SSCFlow employs an iterative denoising mechanism (IDM). The iterative denoising mechanism treats the initialized missing values as corrupted initial imputation and redesigns the MLP as an overcomplete denoising autoencoder (ODA), whereby projecting the latent representation $\dot{z}$ of corrupted initial imputation $\dot{x}$ to a higher dimensional subspace from where the latent representation $\dot{z}$ is then reconstructed as the reconstructed latent representation $\hat{z}$ to be robust for the unreliability of the initialization. The overcomplete representation of denoising autoencoder means that more units exist in successive hidden layers during encoding phase compared to the input layer. The mapping of $\dot{z}$ to a higher dimensional subspace creates representations capable of adding lateral connections, aiding in data recovery \cite{gondara2018mida}. The usefulness of the overcomplete denoising autoencoder is empirically validated in the Sec. \ref{ablation}. The detailed procedure of IDM could be viewed in Fig. \ref{IDM}.

\subsection{Semi-supervised Conditional Network}
\label{Class-Conditional Invertible Transformation}

\begin{figure*}[t]
\centering
\includegraphics[width=1.9\columnwidth]{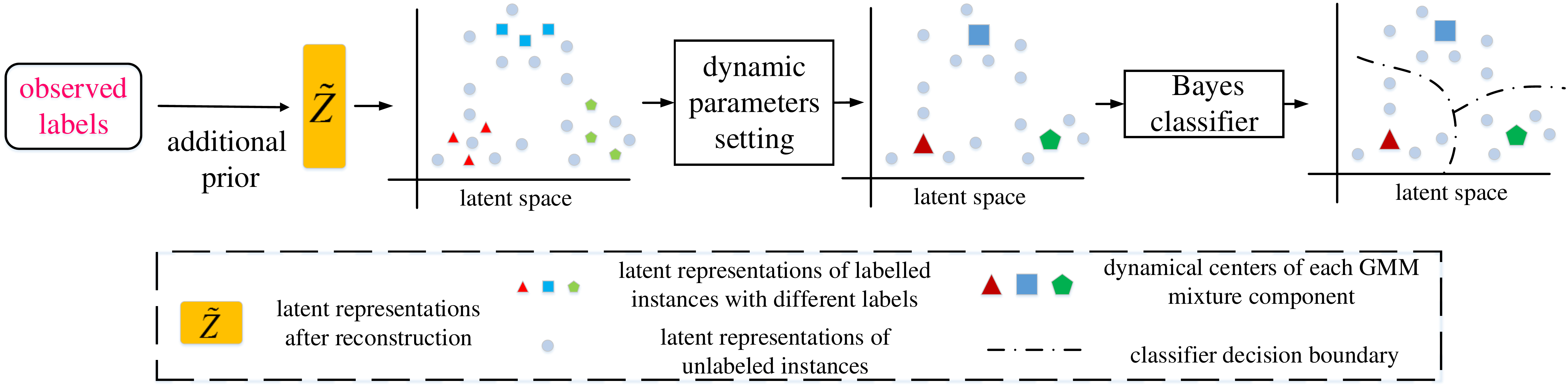} 
\caption{High-level view of class-conditional latent space learning. The observed labels are taken as additional prior knowledges to model the latent space distribution conditioned on the labels as Gaussian mixture model, while the unobserved labels are predicted by maximizing the joint likelihood.}
\label{GMM}
\end{figure*}

In order to incorporate the label information to facilitate the data distribution estimation of the incomplete dataset, SSCFlow needs to estimate the conditional distribution $P_\theta(x|y)$ of labelled and unlabelled instance rather than $P_\theta(x)$. Although there are some existing semi-supervised normalizing flows to estimate conditional distribution with normalizing flows, conditional distribution estimation of unlabelled incomplete instances remains a challenge since it is often estimated indirectly. To take this challenge, SSCFlow deploys a new semi-supervised conditional network, wherein the structure is shared between labelled and unlabelled incomplete instances to facilitate their imputation and classification by directly estimating their conditional distribution. The semi-supervised conditional network enables the direct conditional distribution estimation of unlabelled incomplete instances by treating their unobserved labels as missing values to let them directly participate in the conditional distribution estimation. These unobserved labels are reconstructed and predicted by two modules in semi-supervised conditional network respectively, named semi-supervised conditional invertible transformation and semi-supervised conditional latent space learning.

The semi-supervised conditional invertible transformation explicitly takes the observed labels as additional features to direct the conditional distribution estimation and \textbf{reconstructs} the unobserved labels in the input space of normalizing flow during imputation. In the semi-supervised conditional invertible transformation, the label $y$ is encoded as a one-hot class vector $\dot{y}$ with $\dot{y}^k = 1$ if $y = k$, otherwise $\dot{y}^k = 0$. The one-hot class vector is taken as additional feature of the imputed instance $\dot{x}$ and they are concentrated to form a new input variable $\hat{x} = [\dot{x},\dot{y}]$, which allows to model complex dependencies between data distribution and the class labels. As for these unlabelled instances, their one-hot label vector are treated as missing values and initialized to be reconstructed during IDM. Afterwards, this layer splits the input variable $\hat{x}$ into two non-overlapping parts $\hat{x}_{1}$ , $\hat{x}_{2}$ and applies affine transformation based on the first $\hat{x}_{1}$ to the other $\hat{x}_{2}$. The semi-supervised conditional invertible transformation $T_\theta$ is defined as follow:
\begin{flalign*}
\label{semi-supervised conditional invertible transformation}
&& &1:\dot{y} = OneHotEncoder(y) \\
&& &2:\hat{x} = [\dot{x},\dot{y}]\\
&& &3:\hat{x}_{1},\hat{x}_{2} = Split(\hat{x})\\
&& &4:\dot{z}_{1}=\hat{x}_{1}\\
&& &5:\dot{z}_{2}=\hat{x}_{2} \odot exp(s_\delta(\hat{x}_{1})) +t_\omega(\hat{x}_{1})\\
&& &6:\dot{z} = Concentrate(\dot{z}_{1},\dot{z}_{2}) &&\refstepcounter{equation}\tag{\theequation}
\end{flalign*} where $s$ and $t$ are arbitrary neural networks parameterized by the parameters $\delta$ and $\omega$ with $\theta = \{\delta,\omega\}$. $Split$ denotes the operation that splits the input variable $\hat{x}$ into two non-overlapping parts $\hat{x}_1$ and $\hat{x}_2$, and $Concentrate$ is the inverse operation of $Split$ to concentrate $\dot{z}_{1}$ and $\dot{z}_{2}$ into the latent representation $\dot{z}$. To enhance the performance of the model, steps 3-6 are stacked by using $\dot{z}$ of the $k$-th $\hat{x}^k$ as $\hat{x}^{k+1}$ of the next stack, which follows the design of RealNVP \cite{dinh2017density}. The detailed procedure of the semi-supervised conditional invertible transformation is illustrated in Fig. \ref{affine coupling layer}. As the semi-supervised conditional invertible transformation reconstructs the unobserved label during the imputation of IDM, the unobserved labels always participate in the training process of the unlabelled incomplete instances, which enables direct conditional distribution estimation for the unlabelled incomplete instances. Moreover, the unlabelled instances could also be classified based on the \textbf{reconstructed} class label.


As for the semi-supervised conditional latent space learning, it explicitly takes the observed labels as additional prior knowledges to direct the conditional distribution estimation and assists the \textbf{prediction} of unobserved labels in the latent space of normalizing flow. Traditional normalizing flows model data $x$ as a deterministic and invertible transformation $x=T^{-1}_\theta(z)$ of a simple random latent variable $z$, e.g. standard normal \cite{9089305}, in the latent space, which may not take the additional supervised information into consideration. SSCFlow follows FlowGMM \cite{izmailov2019semisupervised} to take the labels as additional prior knowledges and model the latent space distribution conditioned on the label $k$ as Gaussian with mean $\mu_k$ and covariance $cov_k$:
\begin{flalign*}
\label{Class Conditional Likelihoods}
&& P_Z(z|y=k) = N(z|\mu_k,cov_k) &&\refstepcounter{equation}\tag{\theequation}
\end{flalign*} and the marginal distribution of $z$ is a Gaussian mixture model with the distribution given below:
\begin{flalign*}
\label{marginal likelihoods}
&& P_Z(z) =\sum_{k=1}^{L}N(z|\mu_k,cov_k)P(y=k). &&\refstepcounter{equation}\tag{\theequation}
\end{flalign*} 
Combining Eq. \ref{change of variable theorem} and \ref{marginal likelihoods} , the likelihood for labelled data is 
\begin{flalign*}
\label{labeled likelihoods}
&& P_X(x|y=k) =N(T_\theta(x)|\mu_k,cov_k) |det J_{T_\theta}(x)| &&\refstepcounter{equation}\tag{\theequation}
\end{flalign*}  and the likelihood for data with unobserved label is
\begin{flalign*}
\label{unlabeled likelihoods}
&& P_X(x) & =\sum_{k=1}^{L}P_X(x|y=k)P(y=k)\\
&& &=\sum_{k=1}^{L}N(T_\theta(x)|\mu_k,cov_k) |det J_{T_\theta}(x)|P(y=k). &&\refstepcounter{equation}\tag{\theequation}
\end{flalign*}  The label $y$ of the unlabelled instances could also be \textbf{predicted} by a Bayes classifier, which maximizes the above likelihood with Bayes decision rule according to
\begin{flalign*}
\label{unlabeled prediction}
&& y &= \mathop{\arg\max}_{k}P_X(y=k|x)  \\
&&   &= \mathop{\arg\max}_{k}\frac{P_X(x|y=k)P(y=k)}{P_X(x)} \\
&&   &= \mathop{\arg\max}_{k}P_X(x|y=k)P(y=k). &&\refstepcounter{equation}\tag{\theequation}
\end{flalign*}

Although theoretically the prior distribution should not matter since it could be transformed from the standard normal, the resulting transformations may become easier to learn in practice if additional prior knowledges is provided in the prior distribution \cite{9089305}. In \cite{izmailov2019semisupervised}, the means $\mu$ are randomly sampled from the standard normal distribution $\mu_k \sim N(0,I)$, and the covariance matrices $cov$ are set to identity $\sum_i=I$ for all classes, fixed throughout training. However, since the data is upgraded gradually with the IDM, the fixed setting of the parameters $\mu,cov$ for the Gaussian mixture may not apply to SSCFlow.  Therefore, SSCFlow introduces a dynamic parameters setting algorithm, which dynamically upgrades the parameters along with the IDM, leading to enhanced interpretation. Specifically, after each iteration of IDM, the means $\mu_k$ of class $k$ is set to the mean point of latent representations of labelled data in each class : $\mu_k= (1/n_k)\sum^{n_k}_{m=1}T_\theta(x_k^m)$, where $x^k$ represents labelled data points from class $k$ and $n_k$ is the total number of labelled points in that class. As shown in Sec. \ref{Comparison}, dynamically upgradation of the means $\mu$ results in encoding the learning process with the clustering principle, which makes the classification in SSCFlow more interpretable.  The detailed procedure of class-conditional latent space is illustrated in Fig. \ref{GMM}.

\subsection{Training of SSCFlow}
SSCFlow needs to train two parameters: $\theta$ of semi-supervised conditional invertible transformation $T_\theta$ and $\gamma$ of the overcomplete denoising autoencoder $D_\gamma$. Correspondingly, two novel loss functions to train the parameters of SSCFlow are illustrated in this subsection. The training stage of SSCFlow can be formalized as follows: assume $n$ training samples, $x_i$, $i=0,1,...,n-1$ with their masking vector $M(x_i)$  are available. For each incomplete instance $x_i$, a full training instance  $\dot{x}_i$ is formed by replacing the missing values with its imputed values $\dot{x}_i$ according to its mark vector $M(x_i)$ as:
\begin{flalign*}
\label{imputation}
&& \dot{x}_i = x_i \odot M(x_i) +  \dot{x}_i \odot (1-M(x_i)) &&\refstepcounter{equation}\tag{\theequation}
\end{flalign*} where $\odot$ denotes the Hadamard product. The imputed value $\dot{x}_i$ is randomly initialized at the first iteration and iteratively updated by IDM. The class label $y_i$ of incomplete instance $x_i$ is encoded as one-hot vector $\dot{y}_i$ and the unlabelled instance's class label vector is set to 0.5 for each attribute to indicate the uncertainty of its category. The input feature $\hat{x}_i$ is computed by concentrating the full training instance $\dot{x}_i$ with its one-hot class vector $\dot{y}_i$  according to $\hat{x_i} = [\dot{x}_i,\dot{y}_i]$.  With a full training set constructed, learning the optimal set $\theta^*$ of parameters $\theta$ of the semi-supervised conditional invertible transformation $T_\theta$ is accomplished by directly optimizing the log likelihood as in Eq. \ref{train log likelihood} or minimizing the following loss function $L_{\theta}$ as in Eq. \ref{train theta1},\ref{train theta2}:
\begin{flalign*}
\label{train theta1}
&& L_{\theta} = -\frac{1}{n}\sum_{i=0}^{n-1} (log(|det J_{T_\theta}(\hat{x}_i)|) + log(N(T_\theta(\hat{x}_i)|\mu_k,cov_k)) ) &&\refstepcounter{equation}\tag{\theequation}
\end{flalign*} 
and
\begin{flalign*}
\label{train theta2}
&& L_{\theta} = &-\frac{1}{n}\sum_{i=0}^{n-1} (log(|det J_{T_\theta}(\hat{x}_i)|) && \\ && &+ log(\sum_{k=1}^{L}N(T_\theta(x)|\mu_k,cov_k)P(y=k)) ) &&\refstepcounter{equation}\tag{\theequation}\\
\end{flalign*} for labelled and unlabelled instances respectively. The first cost term in Eq. \ref{train theta1},\ref{train theta2} conditions on the class label in the features aspect introduced in the semi-supervised conditional invertible transformation since the computation of Jacobian relies on the additional class features. The second cost term in Eq. \ref{train theta1},\ref{train theta2} refers to the semi-supervised conditional latent space learning with the Gaussian mixture conditioned on the class labels in the prior aspect. These two terms form the semi-supervised condition structure in SSCFlow.

To train $\gamma$ of the overcomplete denoising autoencoder $D_\gamma$, SSCFlow finds the optimal $\gamma^*$ by minimizing the following loss function $L_{\gamma}$:
\begin{flalign*}
\label{train gamma}
&& L_{\gamma} &= -\frac{1}{n}\sum_{i=0}^{n-1}MSE(x^o_i,\dot{x}^o_i)- \beta_iCEE(y_i,\dot{y}_i) - P_X(\dot{x}_i) &&\refstepcounter{equation}\tag{\theequation}
\end{flalign*} where $x^o_i = x_i \odot M(x_i)$ refers to the observed attributes of $x_i$, $\dot{x}^o_i$ refers to the reconstructed observed attributes of $x_i$, $MSE$ denotes the mean-squared error operator, $\dot{y}_i$ denotes the reconstructed class vector, $CEE$ denotes the cross entropy error operator, $\beta_i = 1$ (resp. 0) when $x_i$ is labelled (resp. unlabelled), and $P_X(\dot{x}_i)$ comes from Eq. \ref{labeled likelihoods},\ref{unlabeled likelihoods} for labelled and unlabelled instances respectively. The first cost term in Eq. \ref{train gamma} encourages $D_\gamma$ to reconstruct the latent representation of $x$ that matches the training sample at the observed attributes $x^o_i$. The second cost term encourages $D_\gamma$ to reconstruct the latent representation to match the observed labels.  The third cost term encourages $D_\gamma$ to reconstruct the latent representation of $x$ with the highest density value according to the current density estimate in the latent space. The pseudocode of the training process is provided in Algorithm. \ref{alg_train}. After training the model, SSCFlow imputes missing values and classifies the unlabelled instances in an unified framework according to Algorithm \ref{alg_impute_classify}.

\begin{algorithm}[htb] 
\renewcommand{\algorithmicrequire}{ \textbf{Input:}} 
\renewcommand{\algorithmicensure}{ \textbf{Output:}}
\caption{ Training Process.} 
\label{alg_train} 
	\begin{algorithmic}[1] 
		\REQUIRE ~~\\ 
		Data:$x_i$, $i=0,1,...,N-1$ with their masking vector $M(x_i)$ and their class label $y_i$\\
		\STATE Randomly initialize the missing values in $x_i$ indicated by $M(x_i)$ to form $\dot{x}_i$.\\
		\STATE Encode class label $y_i$ as one-hot vector with unlabelled instances filled with 0.5 in each attributes to form $\hat{y}_i$.\\
		\STATE $\hat{x}_i = [\dot{x}_i,\dot{y}_i]$\\
		\FOR{$iter = 1$ to $nIterations$}
		\FOR{$n = 1$ to $nEpochs$}
		\STATE Forward Pass:\\
		\STATE $\dot{z}_i = T_\theta(\hat{x}_i)$\\
		\STATE $\hat{z}_i = D_\gamma(\dot{z}_i)$\\
		\STATE $\tilde{x}_i = T^{-1}_\theta(\hat{z}_i)$\\
		\STATE $\dot{z}_i = T_\theta(\tilde{x}_i)$\\
		\STATE Back propagation:\\
		\STATE Compute loss $L_{\theta}$ for labelled and unlabelled instances - Eq. \ref{train theta1},\ref{train theta2}\\

		\STATE Compute loss $L_{\gamma}$ - Eq. \ref{train gamma}\\
		\STATE Update $\theta$ and $\gamma$ by backpropagating loss\\
		\ENDFOR
		\STATE Updating $\hat{x}_i$ with $\tilde{x}_i$ - Eq. \ref{imputation}
		\ENDFOR		
	\end{algorithmic}
\end{algorithm}

\begin{algorithm}[htb] 
\renewcommand{\algorithmicrequire}{ \textbf{Input:}} 
\renewcommand{\algorithmicensure}{ \textbf{Output:}}
\caption{ Imputation and Classification Process.} 
\label{alg_impute_classify} 
	\begin{algorithmic}[1] 
		\REQUIRE ~~\\ 
		Data: Test $x_t$ with its masking vector $M(x_t)$\\
		\STATE Randomly initialize the missing values in $x_t$ indicated by $M(x_t)$ to form $\dot{x}_t$.\\
		\STATE Encode class label of $x_t$ as unlabelled one-hot vector with 0.5 filled in each attributes to form $\hat{y}_t$.\\
		\STATE $\hat{x}_t = [\dot{x}_t,\hat{y}_t]$\\
		\STATE $\dot{z}_t = T_\theta(\hat{x}_t)$\\
		\STATE $\hat{z}_t = D_\gamma(\dot{z}_t)$\\
		\STATE $\tilde{x}_t = T_\theta^{-1}(\hat{z}_t)$\\
		\STATE $[\dot{x}_t,\dot{y}_t] = \tilde{x}_t$\\
		\STATE Impute the missing values of $x_t$ with  $\dot{x}_t$ - Eq. \ref{imputation}\\
		\STATE Classify $x_t$ based on the reconstructed one-hot class vector $\dot{y}_t$ and the Gaussian mixture model $N(\mu_k,cov_k)$ in the latent space by $y_t =  \mathop{\arg\max}_{k} N(\hat{z}_t|\mu_k,cov_k)+\dot{y}_t^k$	\\
	\end{algorithmic}
\end{algorithm}

It is evident in Algorithm \ref{alg_train},\ref{alg_impute_classify} that the computational cost of training SSCFlow is $O(nIterations*nEpochs*n)$, which is consistent with the computational cost of training MCFlow \cite{9157706}. In our training process, the model converged in maximum 10 iterations whereas the epoch at each iteration is a power of 2. When SCCFlow is employed for imputation and classification, the computational cost is relatively lower, i.e. $O(n)$.

In order to verify the robustness and effectiveness of the SSCFlow proposed in this paper, the following experiments are performed on several real-world datasets and comparisons are drawn against five state-of-the-art methods.

\section{Experimental Design}
\label{expr}

\begin{table}[htbp]
	\small
	\centering
	\caption{UCI Datasets Characteristics}
	\begin{tabular}{l|l|l|l} 
	\hline Dataset& Samples &  Attributes  &Classes \\ 
	\hline iris & 150 & 4  & 3\\ 
	\hline parkinsons & 195 & 22  & 2\\ 
	\hline Sonar & 208 & 60  & 2\\ 
	\hline ecoli & 336 & 7  & 8\\ 
	\hline banknote & 1372 & 4  & 2\\ 
	\hline breast& 569 & 32  & 2\\ 
	\hline accent recognition & 329 & 12  & 6\\ 
	\hline wifi localization & 2000 & 7  & 4\\ 
	\hline wine white & 4898 & 11  & 7\\ 
	\hline HTRU2 & 17898 & 8  & 2\\ 
	\hline Sensorless & 58509 & 48  & 11\\ 
	\hline

	\end{tabular}
	\label{table1}%
\end{table}%

\begin{table*}[!htbp]
	\small
	\centering
	\caption{Comparisons of the imputation performance with RMSE \textbf{(lower better)} on different datasets}
	\begin{tabular}{|l||l|l|l|l|l|l|} 
	\hline
	dataset/methods & XGBoost & MICE    & missforest & GAIN  & MCFlow        & SCFlow   \\\hline
    parkinsons &- &$ 0.1588 \pm 0.0028 $&$ 0.1221 \pm 0.0021 $&$ 0.1608 \pm 0.0038 $&$ 0.1183  \pm  0.0020     $&$ \textbf{0.1134}  \pm  0.0027  $\\\hline
    ecoli &- &$ 0.2258 \pm 0.0071 $&$ 0.1878 \pm 0.0022 $&$ 0.1901 \pm 0.0073 $&$ 0.1799  \pm   0.0016      $&$ \textbf{0.1733}  \pm  0.0017  $\\\hline
    accent recognition &- &$ 0.1544 \pm  0.0058 $&$ 0.1383 \pm 0.0016 $&$ 0.1517 \pm 0.0065 $&$ 0.1265  \pm  0.0027 $&$ \textbf{0.1231 } \pm  0.0018  $\\\hline
    iris  &- &$ 0.1772 \pm 0.0004 $&$ 0.1908 \pm 0.0029 $&$ 0.2288 \pm 0.0653 $&$ 0.1661  \pm  0.0116  $&$ \textbf{0.1531}  \pm  0.0100  $\\\hline
    Sonar &- &$ 0.2027 \pm 0.0009 $&$ 0.1682 \pm 0.0011 $&$ 0.3517 \pm 0.0036 $&$ 0.1483  \pm   0.0028     $&$ \textbf{0.1423}  \pm  0.0010  $\\\hline
    wifi localization &- &$ 0.1423 \pm 0.0024 $&$ 0.1422 \pm 0.0025 $&$ 0.1639 \pm 0.0174 $&$ 0.1293  \pm  0.0017 $&$ \textbf{0.1225} \pm 0.0020  $\\\hline
    wine white &- &$ 0.1067 \pm 0.0022 $&$ 0.1084 \pm 0.0002 $&$ 0.1058 \pm 0.0022 $&$ 0.0998  \pm 0.0013  $&$ \textbf{0.0988}  \pm  0.0015  $\\\hline
    banknote &- &$ 0.1740 \pm 0.0004 $&$ 0.1662 \pm 0.0019 $&$ 0.2135 \pm 0.0295 $&$ 0.1835  \pm   0.0067      $&$ \textbf{0.1708}  \pm  0.0055  $\\\hline
    breast &- &$ 0.0973 \pm 0.0121 $&$ 0.0885 \pm 0.0003 $&$ 0.0909 \pm 0.0022 $&$ {0.0780 } \pm  0.0007  $&$ \textbf{0.0772}  \pm  0.0033 $ \\\hline
    HTRU2 &- &$ 0.0874 \pm 0.0011 $&$ {0.0786 \pm 0.0001} $&$ 0.0917 \pm 0.0032 $&$ 0.0654  \pm  0.0014  $&$ \textbf{0.0625}  \pm  0.0011  $\\\hline
    Sensorless &- &$ 0.0572 \pm 0.0004 $&$ 0.0455 \pm 0.0001 $&$ 0.0580 \pm 0.0187 $&$ 0.0357  \pm  0.0003  $&$ \textbf{0.0344} \pm  0.0008  $\\\hline

	\end{tabular}
	\label{table_imputation}%
	\begin{flushleft}
	\footnotesize{ Since XGboost directly classifies the incomplete dataset without imputation, its imputation performance is replaced as -. The lowest RMSE values are highlighted in bold.}
	\end{flushleft}
\end{table*}%

\begin{table*}[!htbp]
	\small
	\centering
	\caption{Comparisons of the classification performance with ACC \textbf{(higher better)} on different datasets}
	\begin{tabular}{|l||l|l|l|l|l|l|} 
	\hline
	dataset/methods & XGBoost & MICE    & missforest & GAIN  & {MCFlow}        & {SSCFlow}   \\\hline 
    parkinsons &$ 0.7007 \pm 0.0476 $&$ 0.7013 \pm 0.0690 $&$ 0.7290 \pm 0.0509 $&$ 0.7151 \pm 0.0602 $&$ 0.7131   \pm 0.0544  $&$ \textbf{0.8024}   \pm 0.0496  $\\\hline 
    ecoli &$ 0.5163 \pm 0.0323 $&$ 0.5789 \pm 0.0402 $&$ 0.6088 \pm 0.0275 $&$ 0.5211 \pm 0.0544 $&$ 0.5488   \pm 0.0790  $&$ \textbf{0.6525}   \pm 0.0110  $\\\hline 
    accent recognition &$ 0.3746 \pm 0.0471 $&$ 0.3883 \pm 0.0804 $&$ 0.3952 \pm 0.0813 $&$ 0.3962 \pm 0.0704 $&$ 0.3944   \pm 0.0996  $&$ \textbf{0.5236}   \pm 0.0207  $\\\hline 
    iris  &$ 0.8033 \pm 0.0431 $&$ 0.8567 \pm 0.0244 $&$ 0.8303 \pm 0.0171 $&$ 0.8107 \pm 0.0771 $&$ 0.8400  \pm 0.0443  $&$ \textbf{0.8683}   \pm 0.0291  $\\\hline 
    Sonar &$ 0.5304 \pm 0.0433 $&$ 0.5360 \pm 0.0763 $&$ 0.5406 \pm 0.0809 $&$ 0.5325 \pm 0.0656 $&$ 0.5431  \pm 0.0765  $&$ \textbf{0.6318}   \pm 0.1177  $\\\hline 
    wifi localization &$ 0.8000 \pm 0.0222 $&$ 0.8120 \pm 0.0128 $&$ {0.8189 \pm 0.0110} $&$ 0.7804 \pm 0.0189 $&$ 0.8084   \pm 0.0621  $&$ \textbf{0.8225}   \pm 0.0708  $\\\hline 
    wine white &$ 0.3568 \pm 0.0115 $&$ 0.3848 \pm 0.0303 $&$ 0.3896 \pm 0.0307 $&$ 0.3869 \pm 0.0297 $&$ 0.3711   \pm 0.0428  $&$ \textbf{0.4886}   \pm 0.0549  $\\\hline 
    banknote &$ {0.6795 \pm 0.0158} $&$ 0.7068 \pm 0.0101 $&$ 0.7306 \pm 0.0168 $&$ 0.6798 \pm 0.0350 $&$ 0.6997   \pm 0.0304  $&$ \textbf{0.7704}   \pm 0.0437  $\\\hline 
    breast &$ 0.9091 \pm 0.0218 $&$ 0.9266 \pm 0.0098 $&$ 0.9301 \pm 0.0102 $&$ 0.9191 \pm 0.0151 $&$ 0.9284  \pm 0.0280  $&$ \textbf{0.9837}   \pm 0.0082  $\\\hline 
    HTRU2 &$ 0.9388 \pm 0.0005 $&$ 0.9419 \pm 0.0018 $&$ 0.9418 \pm 0.0019 $&$ 0.9404 \pm 0.0015 $&$ 0.9431   \pm 0.0022  $&$ \textbf{0.9686}   \pm 0.0010 $ \\\hline 
    Sensorless &$ {0.6314 \pm 0.0029} $&$ 0.6588 \pm 0.0205 $&$ 0.7147 \pm 0.0321 $&$ 0.5826 \pm 0.0401 $&$ 0.6469  \pm 0.0183  $&$ \textbf{0.7983}   \pm 0.0545 $ \\\hline 
    
	\end{tabular}
	\label{table_classification}%
	\begin{flushleft}
	\footnotesize{ The highest ACC values are highlighted in bold.}
	\end{flushleft}
\end{table*}%

In this section, firstly, the datasets used in our experiments are introduced, followed by the competing methods and experimental settings. Afterwards, the comparisons between SSCFlow and other methods are reported. Finally, the ablation experiments of SSCFlow are analysed.
\subsection{Datasets}
The experimental datasets are taken from University of California Irvine (UCI) machine learning repository \cite{asuncion2007uci}. The datasets include iris, parkinsons, Sonar, ecoli, banknote, breast, accent recognition, wifi localization, wine white, HTRU2, Sensorless  \cite{asuncion2007uci}.  The characteristics of these datasets including number of samples, attributes and classes are summarized in Table \ref{table1}.

In order to evaluate the performance of SSCFlow and the competing methods on missing value imputation, missing values are artificially generated for all datasets with a fraction of 50\%. Artificial missing values are generated under the missing mechanism of missing completely at random (MCAR)  \cite{lin2019missing} as in \cite{9157706}. MCAR randomly eliminates some values and sets them missing. That way, the possibility of missing is not influenced by values of any other attributes or the attribute itself.

\begin{figure*}[t]
\centering 
  \subfigure[MICE]{\includegraphics[width=0.6\columnwidth]{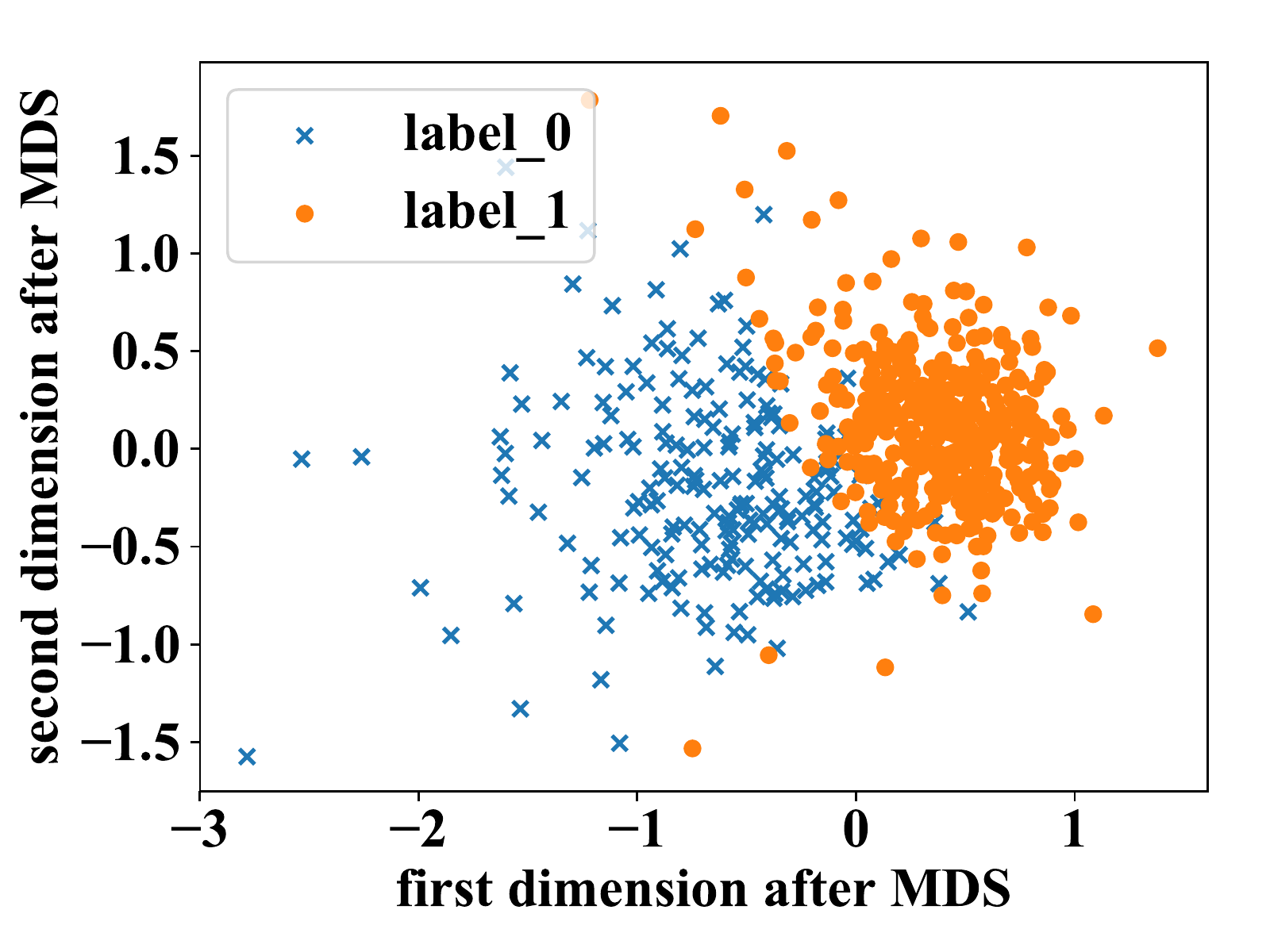}}
  \subfigure[missforest]{\includegraphics[width=0.6\columnwidth]{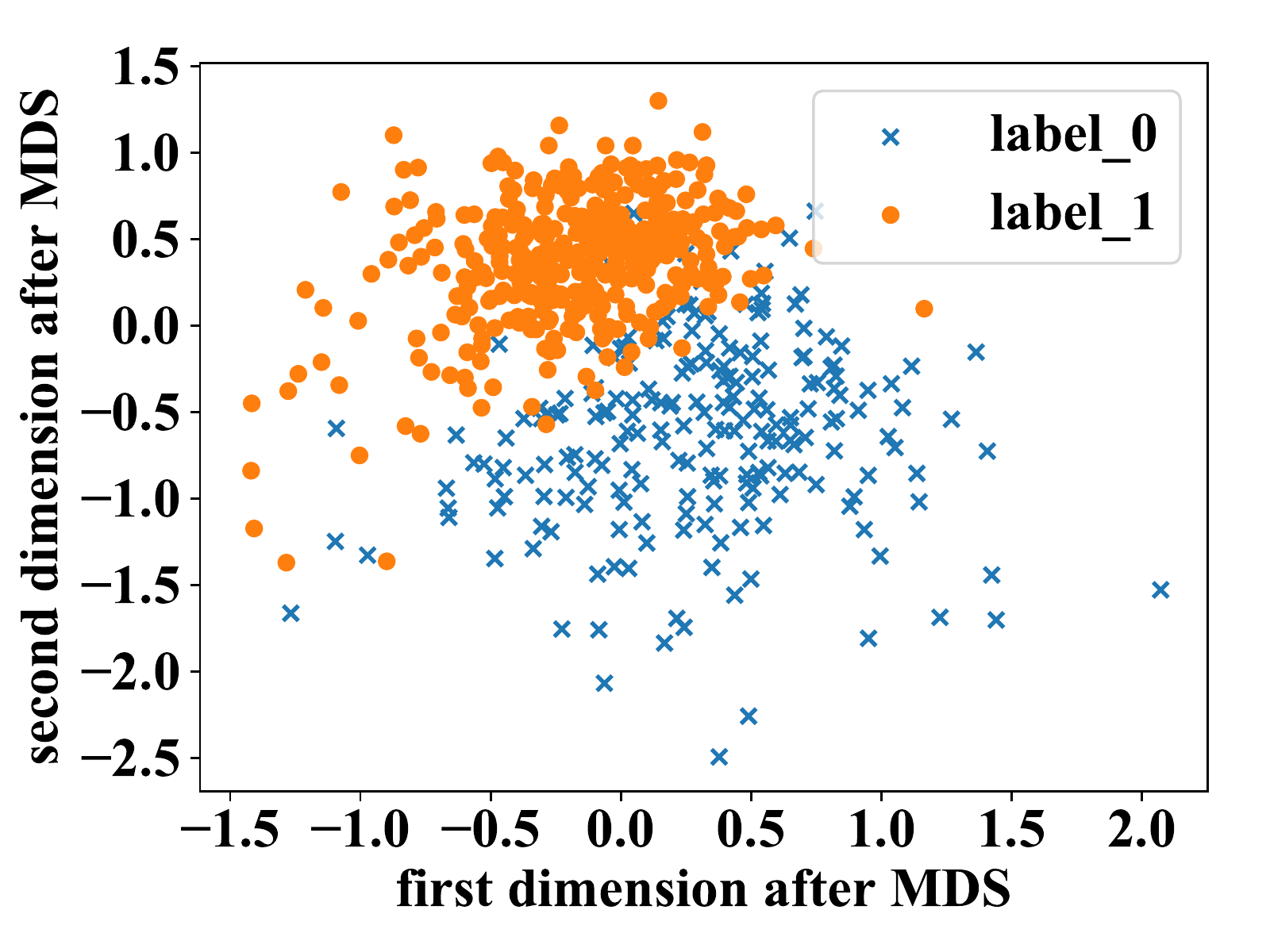}}
  \subfigure[GAIN]{\includegraphics[width=0.6\columnwidth]{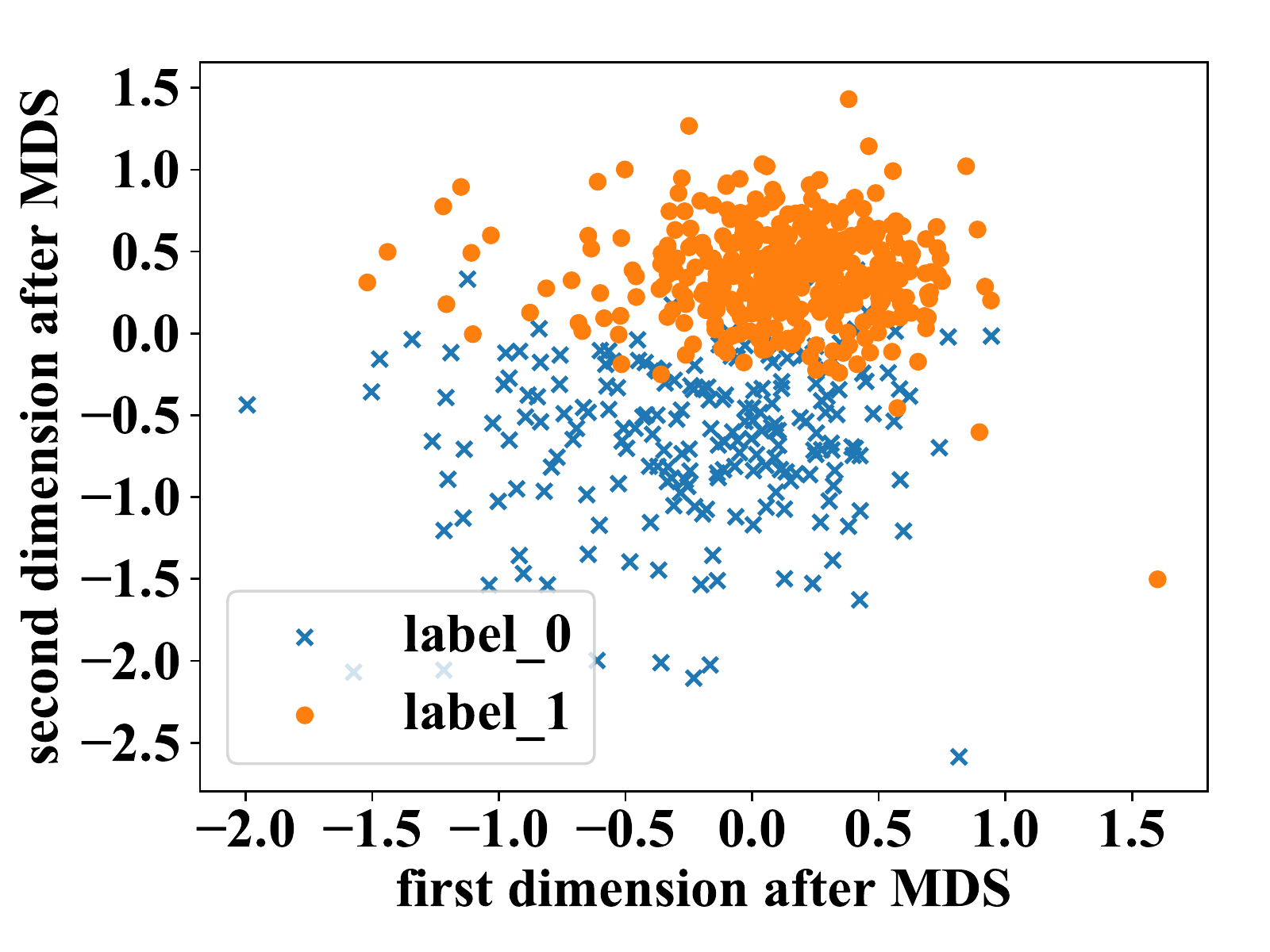}}
  \subfigure[MCFlow]{\includegraphics[width=0.6\columnwidth]{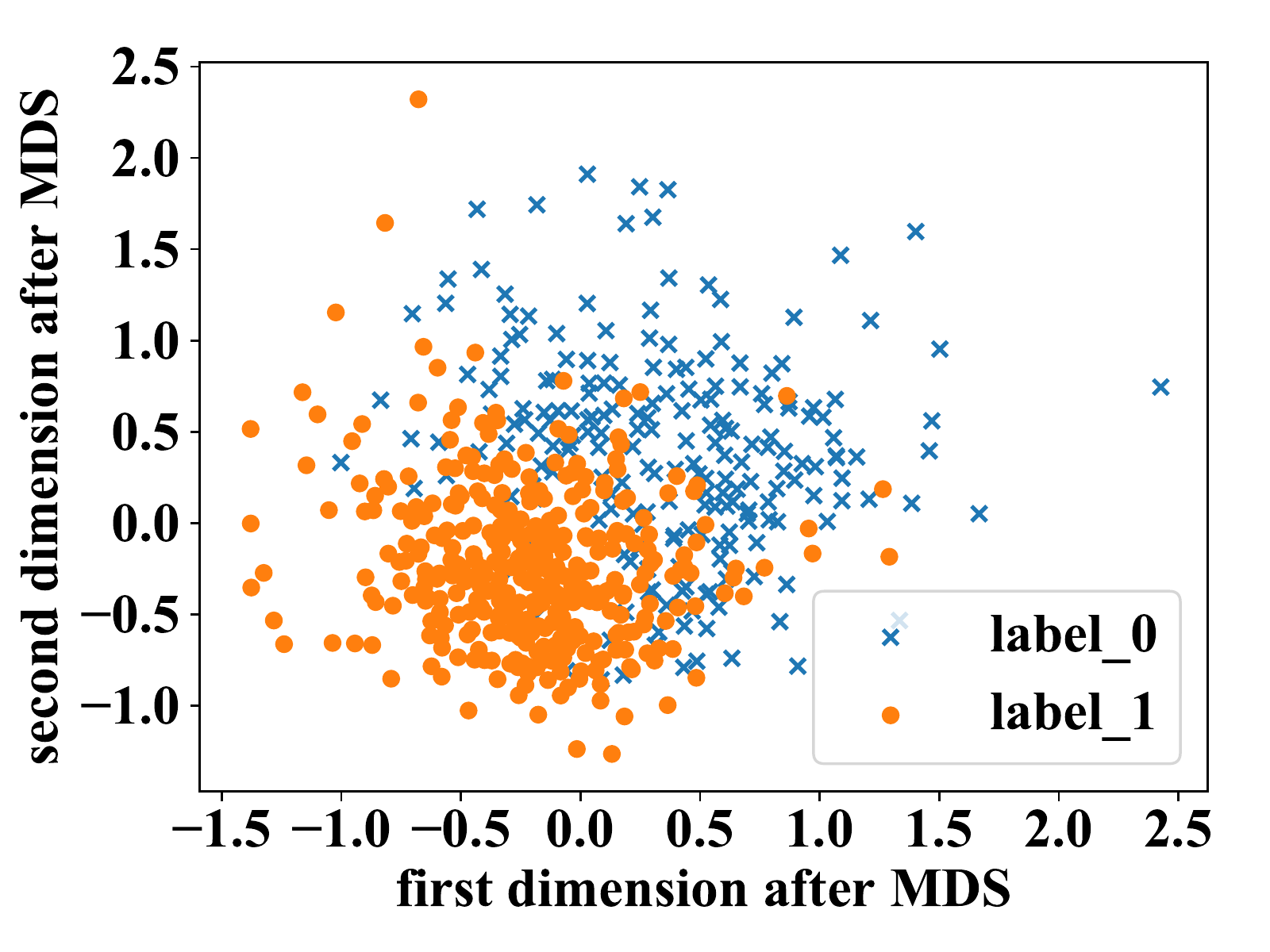}}
  \subfigure[SSCFlow]{\includegraphics[width=0.6\columnwidth]{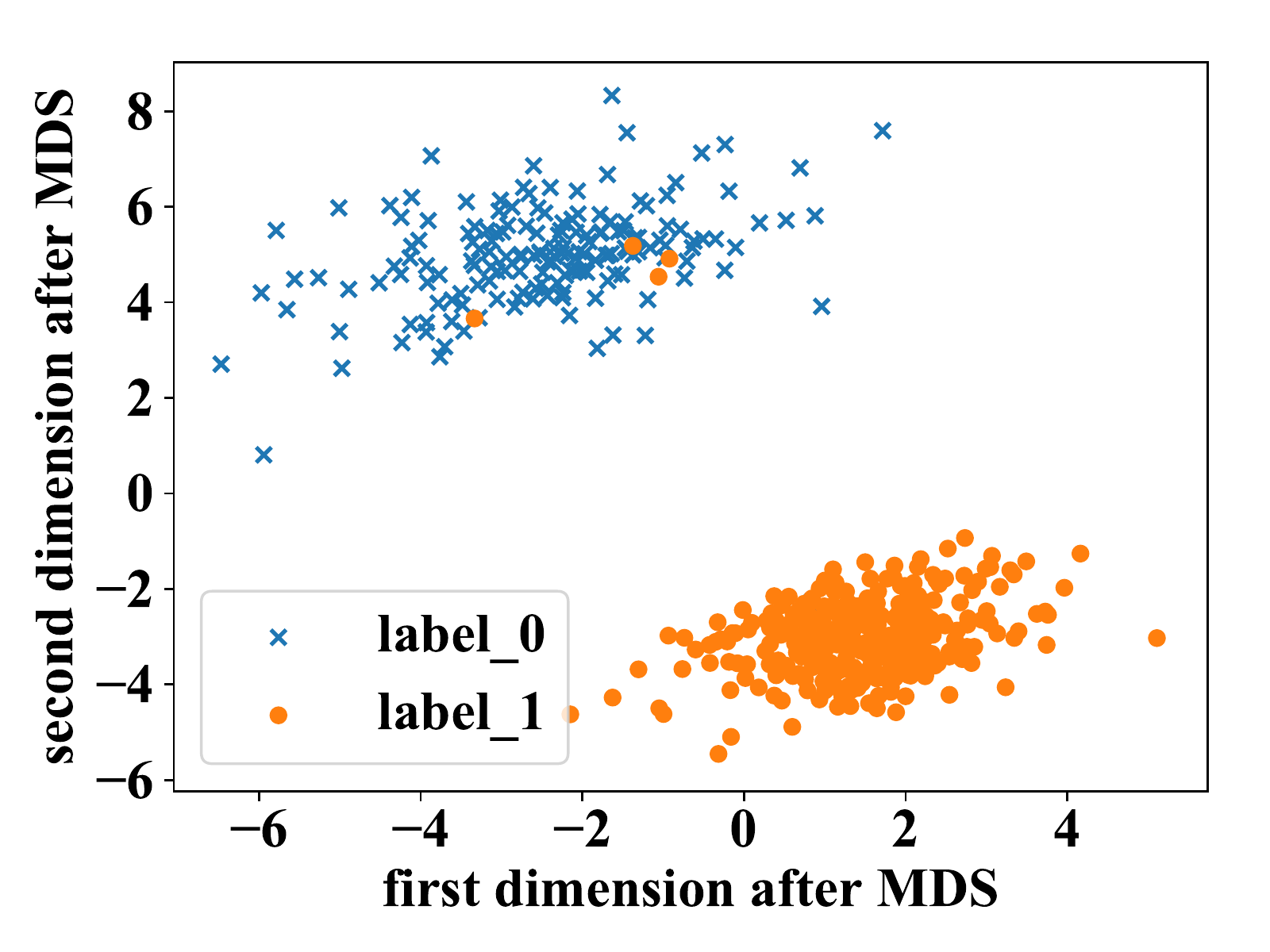}}
\caption{The comparisons between the latent representations of SSCFlow and the imputed dataset of the other MVI methods on dataset \emph{breast}.}
\label{figure_comparison_breast}
\end{figure*}

\begin{figure*}[t]
\centering 
  \subfigure[MICE]{\includegraphics[width=0.6\columnwidth]{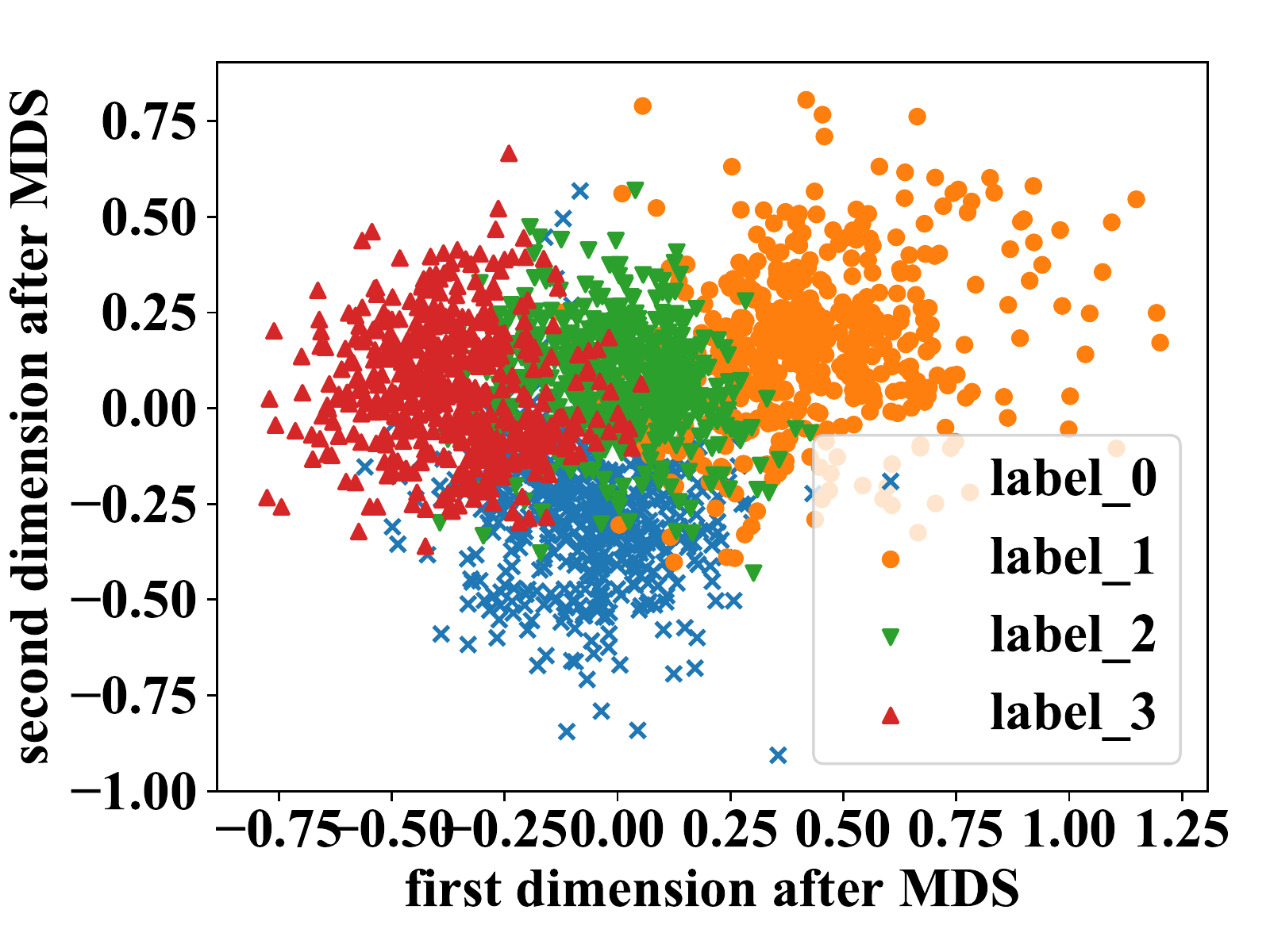}}
  \subfigure[missforest]{\includegraphics[width=0.6\columnwidth]{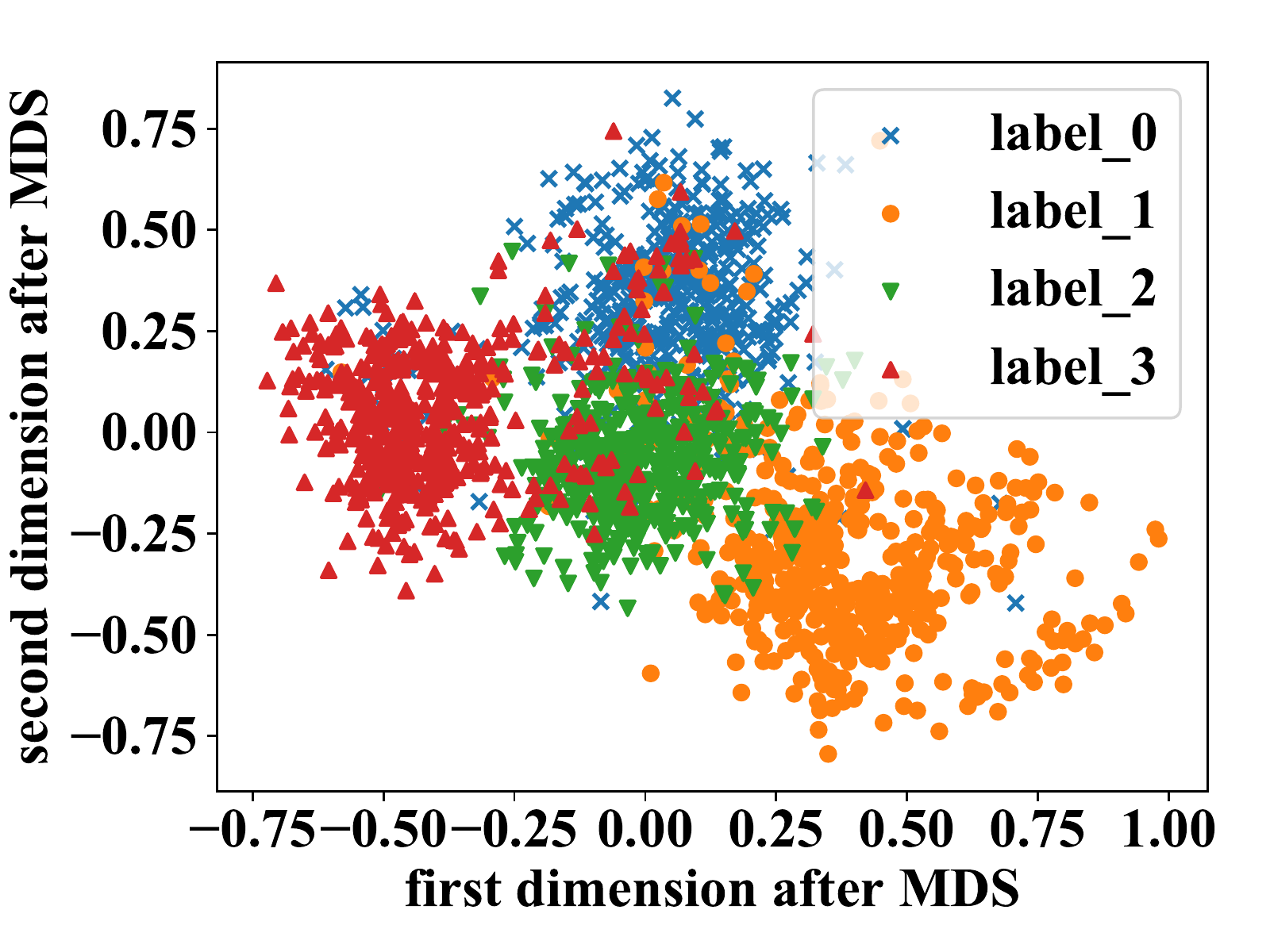}}
  \subfigure[GAIN]{\includegraphics[width=0.6\columnwidth]{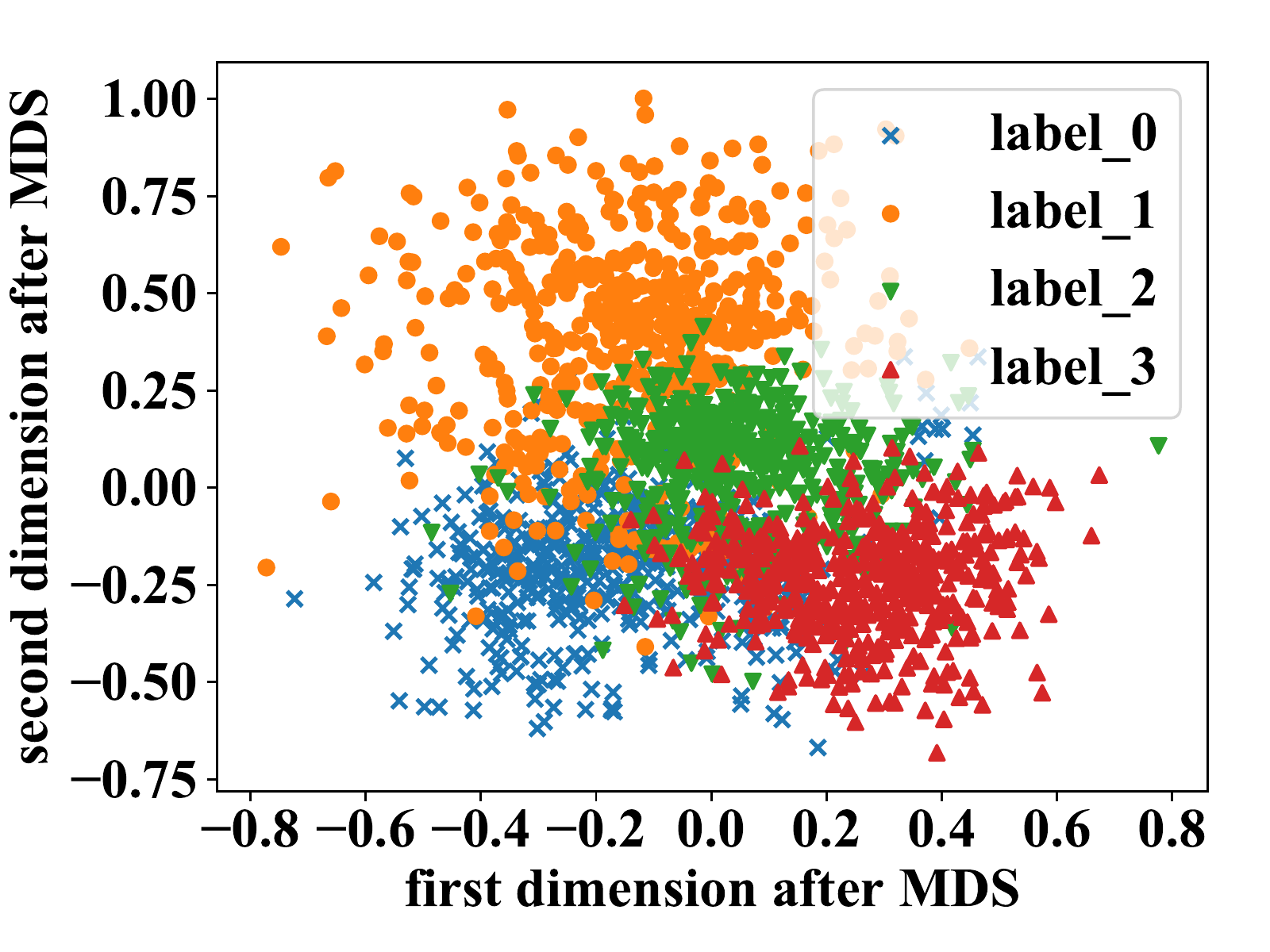}}
  \subfigure[MCFlow]{\includegraphics[width=0.6\columnwidth]{wifi_localization_MCFlow}}
  \subfigure[SSCFlow]{\includegraphics[width=0.6\columnwidth]{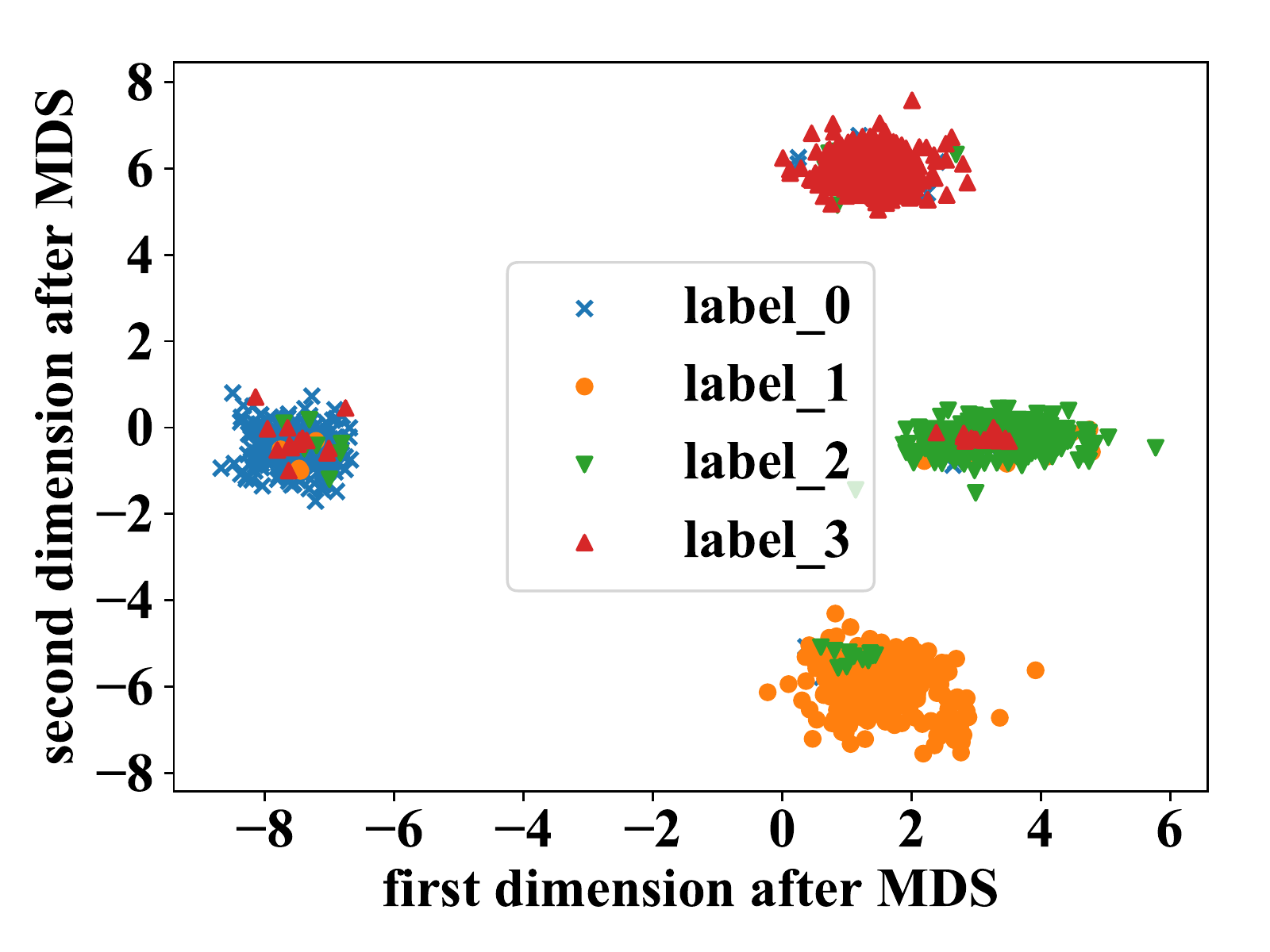}}
\caption{The comparisons between the latent representations of SSCFlow and the imputed dataset of the other MVI methods on dataset \emph{wifi localization}.}
\label{figure_comparison_wifi}
\end{figure*}

\subsection{Comparison Setting}
The proposed SSCFlow is compared with XGBoost\footnote{Implemented as in https://github.com/dmlc/xgboost} \cite{10.1145/2939672.2939785}, missForest\footnote{Implemented as in https://github.com/epsilon-machine/missingpy } \cite{stekhoven2011missforest},  MICE\footnote{Implemented as in https://github.com/iskandr/fancyimpute} \cite{garciarena2017extensive},  GAIN\footnote{Implemented as in https://github.com/jsyoon0823/GAIN} \cite{GAIN}, MCFlow\footnote{Implemented as in https://github.com/trevor-richardson/MCFlow} \cite{9157706} using five-fold cross validation on the incomplete dataset. 

To serve the purpose, three evaluation criteria are used. Firstly, the imputation performance is reported, which is measured using root mean squared error (RMSE) between the missing values introduced into the dataset and the actual values. Afterwards, the comparisons of the classification performance using classification accuracy (ACC) are drawn. The mean and standard deviation of above comparisons result are reported across all folds. Finally, a qualitative comparison is conducted to demonstrate the interpretation superiority of SSCFlow over other methods. In the qualitative comparison, multi-dimensional scaling (MDS) \cite{kruskal1978multidimensional} is employed to reduce the dimensionality of the latent representations in SSCFlow to 2D so that they could be illustrated in figures. The same process is also employed on the imputed dataset of the competing methods to compare them with the latent representations in SSCFlow.

In order to confirm the applicability of SSCFlow as a generative classification algorithm, and facilitate a fair comparison, another generative classification algorithm, implemented as a Bayesian Gaussian mixture, is used after the imputation step of these MVI methods (MICE,  missForest, GAIN and MCFlow). All experiments are implemented using Pycharm 2018 software.

\subsection{Comparison Results and Discussion}
\label{Comparison}
The comparisons of the imputation performance between SSCFlow and the competing methods are illustrated in Table \ref{table_imputation}.  For all datasets, SSCFlow outperforms all other methods with respect to the imputation performance using RMSE. It is observable that SSCFlow reports lower RMSE compared to its base framework MCFlow \cite{9157706} and the other state-of-the-art methods. Moreover, for the datasets where MCFlow does not achieve best imputation performance, SSCFlow scores best imputation performance. It demonstrates that the additional label information introduced by the semi-supervised conditional network helps SSCFlow to better estimate the distribution of missing values.

\begin{table*}[!htbp]
  \centering
  \caption{Ablation Experiments of the imputation performance with RMSE\textbf{(lower better)} on different datasets}
    \begin{tabular}{|l||l|l|l|l|}
    \hline 
    dataset/methods     & {SSCFlow-IDM} &       {SSCFlow-IT} &       {SSCFlow-LS} &      {SSCFlow}   \\\hline 
    parkinsons &$ 0.1143  \pm 0.0015  $&$ 0.1174  \pm 0.0022  $&$ 0.1182  \pm 0.0048  $&$ \textbf{0.1134}  \pm 0.0027$  \\\hline 
    ecoli &$ {0.1738 } \pm 0.0017  $&$ 0.1789  \pm 0.0051  $&$ 0.1795  \pm 0.0073  $&$ \textbf{0.1733}  \pm 0.0017 $ \\\hline 
    accent recognition &$ {0.1261 } \pm 0.0018  $&$ 0.1249  \pm 0.0050  $&$ 0.1272  \pm 0.0017  $&$ \textbf{0.1231}  \pm 0.0018  $\\\hline 
    iris  &$ 0.1620  \pm 0.0035  $&$ 0.1630  \pm 0.0080  $&$ 0.1732  \pm 0.0155  $&$ \textbf{0.1531}  \pm 0.0100 $ \\\hline 
    Sonar &$ 0.1443  \pm 0.0006  $&$ 0.1468  \pm 0.0018  $&$ 0.1478  \pm 0.0013  $&$ \textbf{0.1423}  \pm 0.0010  $\\\hline 
    wifi localization &$ 0.1286  \pm 0.0021  $&$ 0.1261  \pm 0.0012  $&$ 0.1297  \pm 0.0018  $&$ \textbf{0.1225}  \pm 0.0020 $ \\\hline 
    wine white &$ {0.0992 } \pm 0.0006  $&$ 0.1019  \pm 0.0009  $&$ 0.1012  \pm 0.0016  $&$ \textbf{0.0988}  \pm 0.0015 $ \\\hline 
    banknote &$ {0.1714 } \pm 0.0030  $&$ {0.1757 } \pm 0.0060  $&$ 0.1753  \pm 0.0077  $&$ \textbf{0.1708}  \pm 0.0055  $\\\hline 
    breast &$ {0.0780 } \pm 0.0009  $&$ {0.0779 } \pm 0.0012  $&$ {0.0794 } \pm 0.0004  $&$ \textbf{0.0772}  \pm 0.0033 $ \\\hline 
    HTRU2 &$ 0.0800  \pm 0.0015  $&$ 0.0768  \pm 0.0018  $&$ 0.0763  \pm 0.0018  $&$ \textbf{0.0625}  \pm 0.0011 $ \\\hline 
    Sensorless &$ {0.0350 } \pm 0.0007  $&$ {0.0348 } \pm 0.0007  $&$ {0.0354 } \pm 0.0005  $&$ \textbf{0.0344}  \pm 0.0008$  \\\hline 
    \end{tabular}%
  \label{table_ablation_imputation}%
  \begin{flushleft}
	\footnotesize{ The lowest RMSE values are highlighted in bold.}
	\end{flushleft}
\end{table*}%

\begin{table*}[!htbp]
  \centering
  \caption{Ablation Experiments of the classification performance with ACC\textbf{(higher better)} on different dataset}
    \begin{tabular}{|l||l|l|l|l|}
    \hline 
    dataset/methods   & SSCFlow-IDM &        SSCFlow-IT &        SSCFlow-LS &      SSCFlow  \\\hline 
    parkinsons & $0.7258  \pm 0.0485$  & $0.7539  \pm 0.0216$  & $0.6463  \pm 0.1045$  & $\textbf{0.8024}  \pm 0.0496$  \\\hline 
    ecoli & $0.5547  \pm 0.0692$  & $0.5909  \pm 0.1135$  & $0.4361  \pm 0.1307$  & $\textbf{0.6525}  \pm 0.0110$  \\\hline 
    accent recognition & $0.3784  \pm 0.0951$  & $0.4377  \pm 0.1249$  & $0.2538  \pm 0.0098$  & $\textbf{0.5236}  \pm 0.0207$  \\\hline 
    iris  & $0.8467  \pm 0.0267$  & $0.7933  \pm 0.0397$  & $0.8383  \pm 0.0304$  & $\textbf{0.8683}  \pm 0.0291$  \\\hline 
    Sonar & $0.5447  \pm 0.092$2  & $0.5466  \pm 0.0987$  & $0.5695  \pm 0.0945$  & $\textbf{0.6318}  \pm 0.1177$  \\\hline 
    wifi localization & $0.8208  \pm 0.0130$  &$ 0.8120  \pm 0.0178 $ & $0.7889  \pm 0.0199 $ & $\textbf{0.8225}  \pm 0.0708$  \\\hline 
    wine white &$ 0.3739  \pm 0.0370 $ & $0.3850  \pm 0.0244$  & $0.2531  \pm 0.0368$  & $\textbf{0.4886}  \pm 0.0549$  \\\hline 
    banknote & $0.7083  \pm 0.0115$  & $0.7241  \pm 0.0104$  & $0.7303  \pm 0.0178$  & $\textbf{0.7704}  \pm 0.0437$  \\\hline 
    breast & $0.9306  \pm 0.0161$  & $0.9385  \pm 0.0135$  & $0.9354  \pm 0.0176$  & $\textbf{0.9837}  \pm 0.0082$  \\\hline 
    HTRU2 & $0.9453  \pm 0.0018$  & ${0.9602 } \pm 0.0020$  & $0.9516  \pm 0.0127$  & $\textbf{0.9686}  \pm 0.5218$  \\\hline 
    Sensorless & $0.6475  \pm 0.0243$  & $0.6705  \pm 0.1872$  & $0.6002  \pm 0.0107$  & $\textbf{0.7983}  \pm 0.2990$  \\\hline 
    \end{tabular}%
  \label{table_ablation_classification}%
  \begin{flushleft}
	\footnotesize{ The highest ACC values are highlighted in bold.}
	\end{flushleft}
\end{table*}%

The comparisons of classification performance between SSCFlow and the competing methods are illustrated in Table. \ref{table_classification}. SSCFlow yeilds better performance in terms of ACC compared to  MCFlow \cite{9157706} and other state-of-the-art methods. As for the XGBoost \cite{10.1145/2939672.2939785} that could directly classify the incomplete datasets, it achieves the lowest classification accuracy in most of the datasets. It advocates our opinion that ignorance of potential contribution of these missing values may degrade the classification performance. Contrary to this, SSCFlow improves the classification by taking the distribution of missing values into account. A win-win situation is achieved in SSCFlow, as it handles imputation and classification tasks simultaneously.

The comparisons of the latent representations of SSCFlow and the imputed dataset with the other MVI methods in Fig. \ref{figure_comparison_breast},\ref{figure_comparison_wifi} demonstrate the interpretability of SSCFlow. These comparisons are illustrated by MDS in Fig. \ref{figure_comparison_breast},\ref{figure_comparison_wifi}. It's obvious in Fig. \ref{figure_comparison_breast},\ref{figure_comparison_wifi} that the separability of the imputed datasets in the other MVI methods has no significant improvement after the imputation and the biased values from the imputation methods make it even worse. However, it is observable in  Fig. \ref{figure_comparison_breast},\ref{figure_comparison_wifi} that, the classification boundaries of SSCFlow could be clearly determined and naturally encode the clustering principle, which offers better interpretability.  This phenomenon is due to the additional label information, which enables SSCFlow to better estimate the distribution of missing values since the Gaussian mixture model in the latent space guides the latent representation generation with supervised information from the labelled instances. Moreover, it is observable that SSCFlow is able to benefit from the unlabelled data to push the decision boundary to a low-density region during training, as expected. As a result, better classification performance is achieved in SSCFlow with the clear classification boundaries.

\subsection{Ablation Experiments}
\label{ablation}
In this subsection, ablation experiments are reported to demonstrate the improvement provided by the several components of the SSCFlow. The algorithms to be compared in the ablation experiments consist of three parts: SSCFlow-IDM keeps the IDM in SSCFlow, SSCFlow-IT keeps the semi-supervised conditional invertible transformation in SSCFlow, and SSCFlow-LS keeps the semi-supervised conditional latent space learning in SSCFlow. The classification setting in SSCFlow-IDM follows the aforementioned imputation methods with a Bayesian Gaussian mixture classifier. The comparisons result are illustrated in Table \ref{table_ablation_imputation} and \ref{table_ablation_classification}.

It is apparent from Table \ref{table_ablation_imputation} and \ref{table_ablation_classification} that SSCFlow achieves the best performance among these ablation methods, which demonstrates that all components of SSCFlow improve the performance. Moreover, the SSCFlow-LS achieves the worst imputation performance among these methods, which demonstrates that the improvement for imputation would be limited if the label information only participates in the loss computation (density estimation) and does not affect the generative (imputation) process.

\section{Conclusion}
\label{conclusion}
A novel method, SSCFlow, is proposed in this paper for imputation and classification for data with missing values. SSCFlow accomplishes both imputation and classification simultaneously by estimating the conditional probability density of missing values conditioned on their labels. It is conjectured that the conditional probability density estimation could effectively improve the performance of both imputation and classification since the gap between these two tasks is replaced by joint weights optimization and features sharing. Moreover, it is observed that the improvement may be limited when the label information only participates in the loss computation (density estimation) and does not affect the generative (imputation) process.

Even though the proposed method shows promising performance on various incomplete datasets, this paper has some limitations that can be targeted for future research work. For example, the covariance matrices of the Gaussian mixture model in the latent space remain fixed during training, which may limit the expressive ability of the model. The resolution of this issue is intended in the improved version of SSCFlow, in future research.

\bibliographystyle{IEEEtran}
\bibliography{ref}

\begin{thebibliography}{10}
\providecommand{\url}[1]{#1}
\csname url@samestyle\endcsname
\providecommand{\newblock}{\relax}
\providecommand{\bibinfo}[2]{#2}
\providecommand{\BIBentrySTDinterwordspacing}{\spaceskip=0pt\relax}
\providecommand{\BIBentryALTinterwordstretchfactor}{4}
\providecommand{\BIBentryALTinterwordspacing}{\spaceskip=\fontdimen2\font plus
\BIBentryALTinterwordstretchfactor\fontdimen3\font minus
  \fontdimen4\font\relax}
\providecommand{\BIBforeignlanguage}[2]{{%
\expandafter\ifx\csname l@#1\endcsname\relax
\typeout{** WARNING: IEEEtran.bst: No hyphenation pattern has been}%
\typeout{** loaded for the language `#1'. Using the pattern for}%
\typeout{** the default language instead.}%
\else
\language=\csname l@#1\endcsname
\fi
#2}}
\providecommand{\BIBdecl}{\relax}
\BIBdecl

\bibitem{wang2019industrial}
H.~Wang, Z.~Yuan, Y.~Chen, B.~Shen, and A.~Wu, ``An industrial missing values
  processing method based on generating model,'' \emph{Computer Networks}, vol.
  158, pp. 61--68, 2019.

\bibitem{Strawderman89}
W.~E. Strawderman, ``Statistical analysis with missing data (roderick j. a.
  little and donald b. rubin),'' \emph{{Society for Industrial and Applied
  Mathematics} Rev.}, vol.~31, no.~2, pp. 348--349, 1989.

\bibitem{2014Mice}
A.~Hapfelmeier, T.~Hothorn, C.~Riediger, and K.~Ulm, ``Mice: multivariate
  imputation by chained equations in r,'' \emph{International Journal of
  Biostats}, vol.~45, no.~2, pp. 1--67, 2014.

\bibitem{stekhoven2011missforest}
D.~J. Stekhoven and P.~B{\"u}hlmann, ``Missforest — non-parametric missing
  value imputation for mixed-type data,'' \emph{Bioinformatics}, vol.~28,
  no.~1, pp. 112--118, 2011.

\bibitem{9157706}
T.~W. {Richardson}, W.~{Wu}, L.~{Lin}, B.~{Xu}, and E.~A. {Bernal}, ``Mcflow:
  Monte carlo flow models for data imputation,'' in \emph{2020 IEEE/CVF
  Conference on Computer Vision and Pattern Recognition (CVPR)}, 2020, pp.
  14\,193--14\,202.

\bibitem{GAIN}
J.~Yoon, J.~Jordon, and M.~van~der Schaar, ``{GAIN:} missing data imputation
  using generative adversarial nets,'' in \emph{Proceedings of the 35th
  International Conference on Machine Learning (ICML)}, ser. Proceedings of
  Machine Learning Research, vol.~80.\hskip 1em plus 0.5em minus 0.4em\relax
  {PMLR}, 2018, pp. 5675--5684.

\bibitem{dinh2015nice}
L.~Dinh, D.~Krueger, and Y.~Bengio, ``{NICE:} non-linear independent components
  estimation,'' in \emph{3rd International Conference on Learning
  Representations {ICLR}}, 2015.

\bibitem{dinh2017density}
L.~Dinh, J.~Sohl{-}Dickstein, and S.~Bengio, ``Density estimation using real
  {NVP},'' in \emph{5th International Conference on Learning Representations
  (ICLR)}.\hskip 1em plus 0.5em minus 0.4em\relax OpenReview.net, 2017.

\bibitem{10.5555/3327546.3327685}
D.~P. Kingma and P.~Dhariwal, ``Glow: Generative flow with invertible
  1$\times$1 convolutions,'' in \emph{Proceedings of the 32nd International
  Conference on Neural Information Processing Systems (NIPS)}, ser.
  NIPS'18.\hskip 1em plus 0.5em minus 0.4em\relax Red Hook, NY, USA: Curran
  Associates Inc., 2018, p. 10236–10245.

\bibitem{abs-1912-02762}
\BIBentryALTinterwordspacing
G.~Papamakarios, E.~T. Nalisnick, D.~J. Rezende, S.~Mohamed, and
  B.~Lakshminarayanan, ``Normalizing flows for probabilistic modeling and
  inference,'' \emph{CoRR}, vol. abs/1912.02762, 2019. [Online]. Available:
  \url{http://arxiv.org/abs/1912.02762}
\BIBentrySTDinterwordspacing

\bibitem{kruskal1978multidimensional}
J.~B. Kruskal and M.~Wish, \emph{Multidimensional scaling}.\hskip 1em plus
  0.5em minus 0.4em\relax Sage, 1978, vol.~11.

\bibitem{MirzaO14}
\BIBentryALTinterwordspacing
M.~Mirza and S.~Osindero, ``Conditional generative adversarial nets,''
  \emph{CoRR}, vol. abs/1411.1784, 2014. [Online]. Available:
  \url{http://arxiv.org/abs/1411.1784}
\BIBentrySTDinterwordspacing

\bibitem{10.1007/978-3-030-69538-5_37}
J.~Sun, B.~Bhattarai, and T.-K. Kim, ``Matchgan: A self-supervised
  semi-supervised conditional generative adversarial network,'' in
  \emph{Computer Vision -- Asian Conference on Computer Vision (ACCV)
  2020}.\hskip 1em plus 0.5em minus 0.4em\relax Cham: Springer International
  Publishing, 2021, pp. 608--623.

\bibitem{8924906}
S.~Karatsiolis and C.~N. Schizas, ``Conditional generative denoising
  autoencoder,'' \emph{IEEE Transactions on Neural Networks and Learning
  Systems}, vol.~31, no.~10, pp. 4117--4129, 2020.

\bibitem{izmailov2019semisupervised}
P.~Izmailov, P.~Kirichenko, M.~Finzi, and A.~G. Wilson, ``Semi-supervised
  learning with normalizing flows,'' in \emph{Proceedings of the 37th
  International Conference on Machine Learning (ICML)}, ser. Proceedings of
  Machine Learning Research, vol. 119.\hskip 1em plus 0.5em minus 0.4em\relax
  {PMLR}, 2020, pp. 4615--4630.

\bibitem{trippe2018}
\BIBentryALTinterwordspacing
B.~L. Trippe and R.~E. Turner, ``Conditional density estimation with bayesian
  normalising flows,'' \emph{CoRR}, vol. abs/1802.04908, 2018. [Online].
  Available: \url{http://arxiv.org/abs/1802.04908}
\BIBentrySTDinterwordspacing

\bibitem{atanov2020semiconditional}
\BIBentryALTinterwordspacing
A.~Atanov, A.~Volokhova, A.~Ashukha, I.~Sosnovik, and D.~P. Vetrov,
  ``Semi-conditional normalizing flows for semi-supervised learning,''
  \emph{CoRR}, vol. abs/1905.00505, 2019. [Online]. Available:
  \url{http://arxiv.org/abs/1905.00505}
\BIBentrySTDinterwordspacing

\bibitem{YouMDKL20}
J.~You, X.~Ma, Y.~Ding, M.~J. Kochenderfer, and J.~Leskovec, ``Handling missing
  data with graph representation learning,'' in \emph{Advances in Neural
  Information Processing Systems (NIPS)}, vol.~33.\hskip 1em plus 0.5em minus
  0.4em\relax Curran Associates, Inc., 2020, pp. 19\,075--19\,087.

\bibitem{9089305}
I.~{Kobyzev}, S.~{Prince}, and M.~{Brubaker}, ``Normalizing flows: An
  introduction and review of current methods,'' \emph{IEEE Transactions on
  Pattern Analysis and Machine Intelligence}, pp. 1--1, 2020,
  doi:{\color{blue}\href{http://dx.doi.org/10.1109/TPAMI.2020.2992934}{10.1109/TPAMI.2020.2992934}}.

\bibitem{KingmaW13}
D.~P. Kingma and M.~Welling, ``Auto-encoding variational bayes,'' in \emph{2nd
  International Conference on Learning Representations (ICLR)}, 2014.

\bibitem{2014Generative}
I.~Goodfellow, J.~Pouget-Abadie, M.~Mirza, B.~Xu, D.~Warde-Farley, S.~Ozair,
  A.~Courville, and Y.~Bengio, ``Generative adversarial nets,'' in
  \emph{Advances in Neural Information Processing Systems (NIPS)},
  vol.~27.\hskip 1em plus 0.5em minus 0.4em\relax Curran Associates, Inc.,
  2014.

\bibitem{MT}
V.~Bogachev, \emph{Measure Theory}.\hskip 1em plus 0.5em minus 0.4em\relax
  Springer-Verlag Berlin Heidelberg, 2007.

\bibitem{li2020flow}
Y.~Li, S.~Akbar, and J.~Oliva, ``Acflow: Flow models for arbitrary conditional
  likelihoods,'' in \emph{Proceedings of the 37th International Conference on
  Machine Learning (ICML)}, ser. Proceedings of Machine Learning Research, vol.
  119.\hskip 1em plus 0.5em minus 0.4em\relax {PMLR}, 2020, pp. 5831--5841.

\bibitem{6412787}
Y.~Ren, G.~Li, J.~Zhang, and W.~Zhou, ``Lazy collaborative filtering for data
  sets with missing values,'' \emph{IEEE Transactions on Cybernetics}, vol.~43,
  no.~6, pp. 1822--1834, 2013.

\bibitem{6979248}
Y.~Liu, F.~Shang, L.~Jiao, J.~Cheng, and H.~Cheng, ``Trace norm regularized
  candecomp/parafac decomposition with missing data,'' \emph{IEEE Transactions
  on Cybernetics}, vol.~45, no.~11, pp. 2437--2448, 2015.

\bibitem{7959549}
J.~Dai, H.~Hu, Q.~Hu, W.~Huang, N.~Zheng, and L.~Liu, ``Locally linear
  approximation approach for incomplete data,'' \emph{IEEE Transactions on
  Cybernetics}, vol.~48, no.~6, pp. 1720--1732, 2018.

\bibitem{8332950}
Z.~Fang, X.~Yang, L.~Han, and X.~Liu, ``A sequentially truncated higher order
  singular value decomposition-based algorithm for tensor completion,''
  \emph{IEEE Transactions on Cybernetics}, vol.~49, no.~5, pp. 1956--1967,
  2019.

\bibitem{8418828}
Y.~Du, G.~Han, Y.~Quan, Z.~Yu, H.-S. Wong, C.~L.~P. Chen, and J.~Zhang,
  ``Exploiting global low-rank structure and local sparsity nature for tensor
  completion,'' \emph{IEEE Transactions on Cybernetics}, vol.~49, no.~11, pp.
  3898--3910, 2019.

\bibitem{9370000}
E.~Hallaji, R.~Razavi-Far, and M.~Saif, ``Dlin: Deep ladder imputation
  network,'' \emph{IEEE Transactions on Cybernetics}, pp. 1--13, 2021,
  doi:{\color{blue}\href{http://dx.doi.org/10.1109/TCYB.2021.3054878}{10.1109/TCYB.2021.3054878}}.

\bibitem{9407339}
Q.~Shi, Y.-M. Cheung, and J.~Lou, ``Robust tensor svd and recovery with rank
  estimation,'' \emph{IEEE Transactions on Cybernetics}, pp. 1--16, 2021,
  doi:{\color{blue}\href{http://dx.doi.org/10.1109/TCYB.2021.3067676}{10.1109/TCYB.2021.3067676}}.

\bibitem{9418555}
X.~Hu, Y.~Shen, W.~Pedrycz, Y.~Li, and G.~Wu, ``Granular fuzzy rule-based
  modeling with incomplete data representation,'' \emph{IEEE Transactions on
  Cybernetics}, pp. 1--14, 2021,
  doi:{\color{blue}\href{http://dx.doi.org/10.1109/TCYB.2021.3071145}{10.1109/TCYB.2021.3071145}}.

\bibitem{9457222}
F.~Nie, Z.~Li, Z.~Hu, R.~Wang, and X.~Li, ``Robust matrix completion with
  column outliers,'' \emph{IEEE Transactions on Cybernetics}, pp. 1--14, 2021,
  doi:{\color{blue}\href{http://dx.doi.org/10.1109/TCYB.2021.3072896}{10.1109/TCYB.2021.3072896}}.

\bibitem{10.1145/2939672.2939785}
T.~Chen and C.~Guestrin, ``Xgboost: {A} scalable tree boosting system,'' in
  \emph{Proceedings of the 22nd {ACM} {SIGKDD} International Conference on
  Knowledge Discovery and Data Mining, San Francisco, CA, USA, August 13-17,
  2016}.\hskip 1em plus 0.5em minus 0.4em\relax {ACM}, 2016, pp. 785--794.

\bibitem{6848791}
Z.-G. Liu, Q.~Pan, G.~Mercier, and J.~Dezert, ``A new incomplete pattern
  classification method based on evidential reasoning,'' \emph{IEEE
  Transactions on Cybernetics}, vol.~45, no.~4, pp. 635--646, 2015.

\bibitem{8491306}
G.~Wang, J.~Lu, K.-S. Choi, and G.~Zhang, ``A transfer-based additive ls-svm
  classifier for handling missing data,'' \emph{IEEE Transactions on
  Cybernetics}, vol.~50, no.~2, pp. 739--752, 2020.

\bibitem{chen2013learning}
H.~Chen, P.~Ti{\v{n}}o, A.~Rodan, and X.~Yao, ``Learning in the model space for
  cognitive fault diagnosis,'' \emph{IEEE transactions on neural networks and
  learning systems}, vol.~25, no.~1, pp. 124--136, 2013.

\bibitem{10.1145/2487575.2487700}
H.~Chen, F.~Tang, P.~Tino, and X.~Yao, ``Model-based kernel for efficient time
  series analysis,'' in \emph{Proceedings of the 19th ACM SIGKDD International
  Conference on Knowledge Discovery and Data Mining}, ser. KDD '13.\hskip 1em
  plus 0.5em minus 0.4em\relax New York, NY, USA: Association for Computing
  Machinery, 2013, p. 392–400.

\bibitem{2014Cognitive}
H.~Chen, P.~Tino, and X.~Yao, ``Cognitive fault diagnosis in tennessee eastman
  process using learning in the model space,'' \emph{Computers \& Chemical
  Engineering}, vol.~67, no. AUG.4, pp. 33--42, 2014.

\bibitem{chen2015model}
H.~Chen, F.~Tang, P.~Tino, A.~G. Cohn, and X.~Yao, ``Model metric co-learning
  for time series classification.'' in \emph{Proceedings of the Twenty-Fourth
  International Joint Conference on Artificial Intelligence (IJCAI)}.\hskip 1em
  plus 0.5em minus 0.4em\relax AAAI Press, 2015, pp. 3387--3394.

\bibitem{article}
Y.~Li, B.~Jiang, H.~Chen, and X.~Yao, ``Symbolic sequence classification in the
  fractal space,'' \emph{IEEE Transactions on Emerging Topics in Computational
  Intelligence}, vol.~5, no.~2, pp. 168--177, 2021.

\bibitem{2018Multiobjective}
Z.~Gong, H.~Chen, B.~Yuan, and X.~Yao, ``Multiobjective learning in the model
  space for time series classification,'' \emph{IEEE Transactions on
  Cybernetics}, vol.~49, no.~3, pp. 918--932, 2019.

\bibitem{2016Model}
Z.~Gong and H.~Chen, ``Model-based oversampling for imbalanced sequence
  classification,'' in \emph{Proceedings of the 25th {ACM} International
  Conference on Information and Knowledge Management (CIKM), pages =
  {1009--1018}, publisher = {{ACM}}, year = {2016},}.

\bibitem{2017Sequential}
------, ``Sequential data classification by dynamic state warping,''
  \emph{Knowledge and Information Systems}, vol.~57, no.~3, pp. 545--570, 2018.

\bibitem{2019Short}
Y.~Li, J.~Hong, and H.~Chen, ``Short sequence classification through
  discriminable linear dynamical system,'' \emph{IEEE Transactions on Neural
  Networks and Learning Systems}, pp. 3396--3408, 2019.

\bibitem{gondara2018mida}
L.~Gondara and K.~Wang, ``Mida: Multiple imputation using denoising
  autoencoders,'' in \emph{Pacific-Asia Conference on Knowledge Discovery and
  Data Mining}.\hskip 1em plus 0.5em minus 0.4em\relax Springer, 2018, pp.
  260--272.

\bibitem{asuncion2007uci}
A.~Asuncion and D.~Newman, ``Uci machine learning repository,'' 2007.

\bibitem{lin2019missing}
W.-C. Lin and C.-F. Tsai, ``Missing value imputation: a review and analysis of
  the literature (2006--2017),'' \emph{Artificial Intelligence Review}, pp.
  1--23, 2019.

\bibitem{garciarena2017extensive}
U.~Garciarena and R.~Santana, ``An extensive analysis of the interaction
  between missing data types, imputation methods, and supervised classifiers,''
  \emph{Expert Systems with Applications}, vol.~89, pp. 52--65, 2017.

\end{thebibliography}

%








\end{document}